\documentclass[lettersize,journal]{IEEEtran}
\usepackage{amsmath,amsfonts}
\usepackage{algorithmic}
\usepackage{algorithm}
\usepackage{array}
\usepackage[caption=false,font=normalsize,labelfont=sf,textfont=sf]{subfig}
\usepackage{textcomp}
\usepackage{stfloats}
\usepackage{url}
\usepackage{verbatim}
\usepackage{graphicx}
\usepackage{cite}
\usepackage{algorithm}
\usepackage{algorithmic}
\usepackage{balance}
\usepackage{subcaption}

\usepackage{hyperref}
\usepackage{url}
\newtheorem{theorem}{Theorem}[section]

\newtheorem{lemma}[theorem]{Lemma}

\newtheorem{assumption}[theorem]{Assumption}

\usepackage{graphicx} 

\title{Communication-Efficient Zero-Order and First-Order Federated Learning Methods over Wireless Networks}
\author{Mohamad Assaad$^\dag$, Senior, IEEE, Zeinab Nehme$^\dag$, Student, IEEE, Merouane Debbah$^\dag$$^\ddag$, Fellow, IEEE \\ $^\dag$ Laboratory of Signals and Systems (L2S), CentraleSupelec, University of Paris-Saclay, France\\$^\ddag$ Center for 6G Technology, Khalifa University of Science and Technology, UAE \\ Emails:\{Mohamad.Assaad, Zeinab.Nehme, Merouane.Debbah \}@centralesupelec.fr }
\date{August 2024}

\begin{document}
\allowdisplaybreaks
\maketitle

\begin{abstract}

Federated Learning (FL) is an emerging learning framework that enables edge devices to collaboratively train ML models without sharing their local data. FL faces, however, a significant challenge due to the high amount of information that must be exchanged between the devices and the aggregator in the training phase, which can exceed the limited capacity of wireless systems. In this paper,  two communication-efficient FL methods are considered where communication overhead is reduced by communicating scalar values instead of long vectors and by allowing high number of users to send information simultaneously. The first approach employs a zero-order optimization technique with two-point gradient estimator, while the second involves a first-order gradient computation strategy. The novelty lies in leveraging channel information in the learning algorithms, eliminating hence the need for additional resources to acquire channel state information (CSI) and to remove its impact, as well as in  considering asynchronous devices. We provide a rigorous analytical framework for the two methods, deriving convergence guarantees and establishing appropriate performance bounds.

\end{abstract}


\section{Introduction}
\label{intro}
Federated Learning (FL) \cite{FL1} is a privacy-preserving edge AI approach where decentralized clients collaboratively train a shared model. In each round, clients compute gradients based on their local data and send them to a central server. The server aggregates these updates, refines the global model, and redistributes it to all clients. This process repeats until convergence, enabling efficient learning without exposing raw data.
Various first \cite{FL1, FL2, FL3} and second-order \cite{FLS, FLS2}  optimization methods have been proposed to enhance FL performance, but they demand excessive communication and computation resources. With rapidly growing mobile/wireless data \cite{FLchallenges}, FL struggles to efficiently process large-scale data without exhausting system resources. Expanding model sizes further exacerbate this issue by increasing gradient computations and network transmissions of high-dimensional vectors.
In more detail, the high communication overhead in FL comes from two factors: high number of devices and high amount of information to be sent by each device. Over the air computation methods have been considered recently in \cite{FL_wir1,FL_wir3} to allow direct aggregation of the received signals at the receiver. To the best of our knowledge, the proposed solution assumes channel knowledge and are based mainly on analog communication between the devices and the aggregator. This can effectively reduce the communication and computation overhead when the number of devices is high. However, it cannot reduce the amount of information  communicated by each device. On the other hand,  several communication reduction strategies have recently been developed. In \cite{localSGD}, each device performs multiple local updates before sending the local model updates to the server. \cite{pp, FL1} employ selective client participation to limit the number of devices per round. Furthermore, compression techniques, including quantization \cite{CQ1, CQ2, CQ3}, delta encoding \cite{CQ4}, and sparsification \cite{S1} (where only the most significant components of the gradient differences are transmitted) decrease uplink overhead while maintaining model performance. Yet, the quantity of information sent over wireless connections remains substantial despite using these techniques. For example, the number of parameters used typically in small to medium size Neural Network varies between $10^5$ to several millions of parameters. Even with good compression rate, the number of parameters to transmit  in each FL training round is still high, which may exceed the capacity of wireless networks, resulting in a severe delay in FL training. Furthermore, In FL, the information is transmitted over the wireless network and is subject to channel-induced distortions. Devices share the gradient vector \( g \in \mathbb{R}^d \) with the central server, which receives \( H g + n \), where \( H \) represents the channel coefficients and \( n \) is the noise. To obtain the knowledge of the gradient, the unknown channel matrix \( H \) must first be estimated. This process requires resources for transmitting pilot signals to obtain Channel State Information (CSI) and computational resources to decode the gradient from the received signal. 
In this paper, we adopted two approaches, based either Zero-Order (ZO) optimization or on a randomized First-Order (FO) method, to alleviate the communication burden in wireless networks. ZO is   particularly useful when gradient information is unavailable or computationally infeasible, relying instead on function evaluations to estimate gradients \cite{ref2, ref2-ref}. 
 
 In this paper, we consider a federated optimization problem where the training is performed using two  methods: either a zeroth-order (ZO) gradient estimation approach or a first-order gradient-based method. To achieve communication efficient learning, both of methods involve only sending two single scalars per communication round over the wireless network, instead of the typically  high dimensional gradient vectors used in FL. The novelty lies in incorporating the channel effects directly into the learning algorithms, avoiding then the need to estimate the channel at channel matrix H and to remove its impact at the receiver. The phase shift introduced by the channel is not removed and treated as an additional perturbation in the model. The learning methods consist in sending two scalars by each device in two successive minislots. The first one contains a fixed predefined symbol and the second one contains the ZO or FO gradient estimate. By aggregating the received symbols from all devices in each minislot and then making the product of the two aggregated scalars, we show that the aggregator is able to make a biased or unbiased estimate of the gradient. We then establish theoretically the convergence of both algorithms in nonconvex settings. It is worth mentioning that handling  nonconvexity is inherently difficult, as it complicates both optimization progress and convergence analysis.  In our work, the convergence is established by taking into account the channel disturbance that is not removed at the receiver as explained above. More importantly, the considered learning methods, along with the theoretical convergence analysis, are extended to the case where the devices are asynchronous. Furthermore, we provide a sample path convergence guarantee with high probability, making thus our work different from prior work in the literature.

In \cite{allerton, ElissaJMLR24}, zeroth-order (ZO) optimization under analog communication assumptions is investigated. However, that work considers a different algorithm necessitating two rounds of communication between the devices and the aggregator to make one biased gradient estimate, where in each round a vector of size equivalent to the model size is communicated in the downlink. In our work here, only one round is needed to make a gradient estimate, and only two scalars are sent in successive minislots in the uplink. Furthermore, we consider  ZO and FO-based FL methods and establish the convergence for synchronous and asynchronous devices, while  \cite{allerton,ElissaJMLR24} is limited to a ZO approach with synchronous devices.  In \cite{ElissaTSP24}, the impact of quantization on ZO in an FL setting has been investigated. However, the work in \cite{ElissaTSP24} is limited to ZO methods, synchronized devices, and under the assumption that each device is transmitting on a different resource and the channel is known perfectly. For a large number of devices, the approach in \cite{ElissaTSP24}  requires hence large amount of resources. In this paper, however, all devices are transmitting on the same resource, resulting thus in a huge reduction in communication overhead. In addition, both  FO and ZO methods with low communication overhead are considered, and the analysis is made for synchronous and asynchronous devices, along  with a high probability sample path convergence guarantee analysis.  
Recent work in \cite{malladi} has investigated the use of zeroth-order (ZO) methods in large language models (LLMs) to reduce computational costs during fine-tuning. The authors provide a regret-bound analysis in a centralized setting, assuming the loss function satisfies the Polyak-Łojasiewicz (PL) condition. In contrast, our paper considers  a federated learning (FL) setting with two algorithms (ZO and FO) that incorporate the channel disturbance in the learning. In addition, our proof is done for general nonconvex settings. Additionally, prior studies \cite{DistLLM23,FLforward24} have examined the potential of ZO methods to reduce communication and memory overhead in distributed LLM training. However, unlike our work, these studies consider a standard ZO method and do not offer a formal convergence analysis.
\cite{FedZO} proposes a federated zero-order optimization method for wireless networks. However, their approach is essentially a direct adaptation of first-order methods to the zero-order setting: devices perform local updates using ZO gradient estimates and transmit the full model vector to the central server. To improve communication efficiency, the authors employ multiple local stochastic gradient-like steps with ZO estimates and partial device participation per round. Their analysis derives a gradient decay bound over a finite horizon T under a fixed stepsize, which itself depends on T. In our work, two different ZO and FO methods have been proposed where only two scalars are exchanged by each device, the channel disturbance is incorporated in the learning and with the use of over the air aggregation.  \cite{Fedscalar} reduces uplink communication overhead per device but ignores wireless channel effects, assumes synchronous devices and dedicated resources per device, and requires local gradient computations. In this paper, we consider both ZO and FO based methods (ZO does not require gradient computation), integrate the wireless channel perturbation in the learning method, and consider the asynchronous device scenario.

\section{System Model and FL algorithms}
\subsection{System Model}
Consider an FL framework consisting of $N$ edge devices that locally train models using their private datasets and a central server coordinating with these edge devices to train a global model $\theta \in \mathbb{R}^d$ over a wireless network. Let $\mathcal{N}=\{1,...,N\}$ be the set of devices and $F_i:\mathbb{R}^d\rightarrow\mathbb{R}$ be the loss function associated with the local data stored on device $i$, $\forall i\in\mathcal{N}$. The goal is to minimize the global loss function:
\begin{align}
    F(\theta) = \sum_{i=1}^{N} F_i(\theta), \quad \text{where} \quad F_i(\theta) = \mathbb{E}_{\xi_i \sim D_i} [f_i(\theta, \xi_i)]\label{F_i}
\end{align}
where $\xi_i$ is an i.i.d. ergodic stochastic process following a local data distribution $D_i$. The work includes the case of nonconvex functions $F$, $F_i$, and $f_i$. In standard FL methods, each device updates the model locally by computing the gradient of the loss function $F_i$ and then uploading their local gradients or models to the server. This requires computation and high communication overhead since the gradient is a long vector of size $d$. In this paper, only two scalars are communicated by each device. 

Each device communicates with the central server over a wireless channel subject to fading and noise. The channel coefficient between device $i$ and the server at time slot $k$ is denoted as $\tilde{h}_{i,k}$, which is assumed to be a zero-mean random variable with standard deviation $\sigma_h$. 
 The received signal at the server from device \( i \) at time slot \( k \) is given by :
    \begin{align}
        y_{i,k} = \tilde{h}_{i,k} x_{i,k} + \tilde{n}_{i,k},
    \end{align}
    where, \( x_{i,k} \) is the transmitted signal, \( \tilde{h}_{i,k} \) is the channel coefficient, and \( \tilde{n}_{i,k} \sim \mathcal{N}(0, \sigma_n^2) \) represents additive Gaussian noise.

\subsection{Algorithm 1}
This section provides an efficient zero-order federated learning (EZOFL) method that extends previous ZO methods by including the channel coefficient in the learning process. 
At each iteration, every user $i\in\mathcal{N}$ computes two queries of its loss function using its local data and then computes their difference. This provides a two-point estimate for the gradient given by \cite{ref2, ref2-ref}:
\begin{equation}
    \Delta f_{i,k} = f_i\big(\theta_k + \gamma_k\Phi_k, \xi_{i,k}\big)-f_i\Big(\theta_k - \gamma_k\Phi_k, \xi_{i,k}\big)
\end{equation}
    where $\Phi_k$ is a perturbation direction generated randomly and pre-stored in the devices, $\gamma_k$ is the step size that will be specified later in the paper, and  $\theta_k$ is the model at iteration $k$.

At each iteration $k$, every user $i$  transmits in the first minislot one scalar at $a_i$,  a predefined constant  that is set to $a_i = \frac{1}{\mathbb{E}[h_i]^2}$.
    Since large-scale fading changes slowly over time, \(\mathbb{E}[h_i]^2\) can be reliably estimated making it relatively easy to obtain in practice. All users transmits in parallel resulting in the received signal $\sum_{i \in N} a_i \tilde{h}_{i,k} + \tilde{n}_{1,k}$. The server takes the real part of this signal, i.e. $\sum_{i \in N} a_i h_{i,k} + n_{1,k}$.
In the second minislot, each device computes the scalar $\Delta f_{i,k} $ to be sent to the central server which receives $\sum_{i \in N} \Delta f_{i,k} \tilde{h}_{i,k} + \tilde{n}_{2,k}$. The server computes simply the real part of the received signal, which is $\sum_{i \in N} \Delta f_{i,k} h_{i,k} + n_{2,k}$. 
   
The server then computes the product:
\begin{equation}
    \label{prod}
    \left( \sum_{i \in N} a_i h_{i,k} + n_{1,k} \right) \times \left( \sum_{i \in N} \Delta f_{i,k} h_{i,k} + n_{2,k} \right)
\end{equation}
which is broadcast to all devices. Each device reconstructs the proposed  zero-order gradient estimate:
\begin{equation}
\label{grad}
    g_k= \Phi_k  \left( \sum_{i \in N} a_i h_{i,k} + n_{1,k} \right)  \left( \sum_{i \in N} \Delta f_{i,k} h_{i,k} + n_{2,k} \right)
\end{equation}
Finally, each device updates its model using:
\begin{equation}
    \theta_{k+1} = \theta_k - \eta_k g_k
\end{equation}
where $\eta_k$ is the step size. One can see that the channel perturbation (including the phase shift impact of the channel) is included in the gradient estimation.

This method replaces the idea of transmitting long vectors across the network with only two scalars in two successive minislots. This requires very low communication overhead and negligible transmission time. 

\begin{algorithm}[h]
	\caption{The EZOFL Algorithm}
	\label{alg1}
	{\bfseries Input:} initial values: $\theta_0$, $\eta_0$ and $\gamma_0$ 
	\begin{algorithmic}[1]
		\FOR{$k=0, \ldots, K$}
		\STATE $\theta_k$ is the model at iteration $k$. Each device  $i$ computes  $\Delta f_{i,k}=f_i\big(\theta_k + \gamma_k\Phi_k, \xi_{i,k}\big)-f_i\Big(\theta_k - \gamma_k\Phi_k, \xi_{i,k}\big)$ using its local data.
		\STATE The server receives:
        At minislot one:
       $\sum_{i \in \mathcal{N}} a_i \tilde{h}_{i,k} + \tilde{n}_{1,k}$\\
        At minislot two:
        $\sum_{i \in N} \Delta f_{i,k} \tilde{h}_{i,k} + \tilde{n}_{2,k}$
		\STATE The server computes the product given in (\ref{prod}) which is broadcast to all devices
        \STATE Each device multiplies the received product by $\Phi_k$ to obtain $\hat{g}_k$ given in (\ref{grad})
		\STATE Each device updates the model  $\theta_{k+1} = \theta_k - \eta_k g_k$\\
		\ENDFOR 
	\end{algorithmic}
\end{algorithm}

 \subsection{Algorithm 2}
The following algorithm is an extension of \cite{Fedscalar} by also including the channel impact as a random perturbation in the gradient estimation.  Contrary to algorithm 1, this algorithm requites local computation of the gradient by each device. 

At each iteration \( k \), every user \( i \in \mathcal{N} \) computes the gradient of its loss function, denoted as \( \nabla f_{i,k}(\theta_k) \), and then calculates the scalar \( \left( \nabla f_{i,k}(\theta_k) \right)^T \Phi_k \) as in \cite{Fedscalar}, where \( \Phi_k \) is a perturbation direction randomly generated and pre-stored on each device, and \( \theta_k \) is the model at iteration \( k \).

Then, the algorithm proceeds similarly to Algorithm 1. Specifically, each user \( i \) transmits a scalar \( a_i \) in the first minislot, and in the second minislot the user transmits the scalar \( \left( \nabla f_{i,k}(\theta_k) \right)^T \Phi_k \). 
The base station (or server) then receives the following signals:
\[
\sum_{i \in \mathcal{N}} a_i \tilde{h}_{i,k} + \tilde{n}_{1,k} \quad \text{and} \quad \sum_{i \in \mathcal{N}} \left( \nabla f_{i,k}(\theta_k) \right)^T \Phi_k \tilde{h}_{i,k} + \tilde{n}_{2,k},
\]
and, as in Algorithm 1, it computes the real part of each of these signals, that is  $\sum_{i \in \mathcal{N}} a_i h_{i,k} + n_{1,k} $ and $\sum_{i \in \mathcal{N}} \left( \nabla f_{i,k}(\theta_k) \right)^T \Phi_k h_{i,k} + n_{2,k}$, and then it computes the product of the real parts of these two received signals and sends the resulting scalar back to the devices. 
Every user is then able computes the gradient estimate \( \hat{g}_k \) as:
\begin{equation}
    \label{grad2}
g'_k = \Phi_k  ( \sum_{i \in \mathcal{N}} a_i h_{i,k} + n_{1,k} ) ( \sum_{i \in \mathcal{N}} \left( \nabla f_{i,k}(\theta_k) \right)^T \Phi_k h_{i,k} + n_{2,k} ).
\end{equation}
Finally, the model is updated by each device using the following update rule:
\[
\theta_{k+1} = \theta_k - \eta_k g'_k,
\]
where \( \eta_k \) is the step size, specified in Section III.
\begin{algorithm}[h]
	\caption{The EFOFL Algorithm}
	\label{alg:example_2p}
	{\bfseries Input:} initial values of \( \theta_0 \), \( \eta_0 \)  
	\begin{algorithmic}[1]
		\FOR{$k = 0, \ldots, K$}
		\STATE Let \( \theta_k \) be the model at iteration \( k \). Each device \( i \) computes \( \left( \nabla f_{i,k}(\theta_k) \right)^T \Phi_k \) using its local data.
		\STATE The server receives:
		At  minislot one:
		$\sum_{i \in \mathcal{N}} a_i \tilde{h}_{i,k} + \tilde{n}_{1,k}$\\
		At  minislot two:
		$
		\sum_{i \in \mathcal{N}} \left( \nabla f_{i,k}(\theta_k) \right)^T \Phi_k \tilde{h}_{i,k} + \tilde{n}_{2,k}
		$
		\STATE The server computes the product of the two received symbols and broadcasts the result to all devices.
		\STATE Each device multiplies the received product by \( \Phi_k \) to obtain the gradient estimate \( g'_k \) as given in (\ref{grad2}).
		\STATE Each device updates the model: 
		$
		\theta_{k+1} = \theta_k - \eta_k g'_k
		$
		\ENDFOR
	\end{algorithmic}
\end{algorithm}

\section{Convergence analysis}
The following provides the analysis of the behavior of both algorithms in the nonconvex setting. 
Assuming that a global minimizer $\theta^*\in\mathbb{R}^{d}$ exists such that $\min_{\theta\in\mathbb{R}^d} F(\theta) = F(\theta^*)>-\infty$ and $\nabla F(\theta^*)=0$. \\
We start by introducing necessary assumptions on the global objective function:

\begin{assumption}\label{objective_fct} 
 $\nabla F_i(\theta)$ and $\nabla^2 F_i(\theta)$ exist and are continuous. We also assume there exists a constant $b>0$ such that
	$\|\nabla^2 F_i(\theta)\|_2\leq b$,$\forall i\in\mathcal{N}$.	
\end{assumption}
By this Assumption, the objective function $\theta\longmapsto F(\theta)$ is $L$-smooth for some positive constant $L$,
	$\|\nabla F(\theta)-\nabla F(\theta')\|\leq L\|\theta-\theta'\|, \;\forall \theta,\theta'\in\mathbb{R}^d,$	or equivalently, 
	$F(\theta)\leq F(\theta')+\langle\nabla F(\theta'), \theta-\theta'\rangle +\frac{L}{2}\|\theta-\theta'\|^2$.
\begin{assumption}\label{local_fcts}
	All loss functions $\theta\mapsto f_i(\theta,\xi_i)$ are Lipschitz continuous with Lipschitz constant $L_{\xi_i}$,
	$|f_i(\theta,\xi_i)-f_i(\theta',\xi_i)|\leq L_{\xi_i}\|\theta-\theta'\|$, $\forall i\in\mathcal{N}$.
	In addition, $\mathbb{E}_{\xi_i} f_i (\theta, \xi_i) < \infty, \forall i \in\mathcal{N}$.
\end{assumption}
We next consider standard assumptions about the step sizes: 
\begin{assumption}\label{step_sizes_1}
	Both the step sizes $\alpha_k$ and $\gamma_k$ vanish to zero as $k\rightarrow\infty$ and the following series composed of them satisfy the convergence assumptions
	$\sum_{k=0}^{\infty}\alpha_k \gamma_k = \infty$, $\sum_{k=0}^{\infty}\alpha_k\gamma_k^3 <\infty$, and $\sum_{k=0}^{\infty}\alpha_k^2\gamma_k^2<\infty$.
\end{assumption}
An example of step sizes satisfying Assumption \ref{step_sizes_1}: $\alpha_k = \alpha_0(1+k)^{-\upsilon_1}$ and $\gamma_k = \gamma_0 (1+k)^{-\upsilon_2}$ with $\upsilon_1, \upsilon_2>0$. Then, it is sufficient to find $\upsilon_1$ and $\upsilon_2$ such that $0<\upsilon_1+\upsilon_2\leq 1$, $\upsilon_1+3\upsilon_2>1$, and $\upsilon_1+\upsilon_2>0.5$.
\begin{assumption}\label{perturbation}
	$\Phi_{k} = (\phi_{k}^1, \phi_{k}^2, \ldots, \phi_{k}^d)^T$, are generated independently at each iteration, and with i.i.d. elements, that is  $\mathbb{E}(\phi_{k}^{d_1} \phi_{k}^{d_2}) =0$ for $d_1 \neq d_2$. Furthermore,  there exists $b_1 >0$ and $b_2 >0$ such that
	$\mathbb{E} (\phi_{k}^{d_j})^2 = b_1$, $\forall {d_j},k$ and 
	$\|\Phi_{k}\|\leq b_2$, $\forall k$.
\end{assumption}
As an example satisfying  Assumption \ref{perturbation}, one can randomly select every dimension of $\Phi_{k}$ from $\{-1, 1\}$ with equal probability. Then, $b_1=1$ and $b_2=\sqrt{d}$.

\subsection{Convergence of Algorithm 1}
We now provide the convergence of algorithm 1 in the following theorem. 
\begin{theorem}\label{th-ncvx}
	When Assumptions \ref{objective_fct}-\ref{perturbation} hold, we have  $\lim_{k\rightarrow\infty}\mathbb{E}[\|\nabla F(\theta_k)\|^2]=0$  under Algorithm 1.
	
	Proof: Refer to Appendices A-A, A-B, and A-C.. 
\end{theorem}
In the following theorem, we also provide  the achieved convergence rate, which is of the order $O(\frac{1}{\sqrt{K}})$. This rate competes with standard gradient methods while requiring much less communication overhead (two scalars) per iteration instead of a long vector of gradients to exchange per iteration. More importantly, and contrary to classical convergence analysis that is limited to show convergence in expectation, we provide here a high probability sample path convergence. More specifically, we show that after $K$ iterations, each run of Algorithm 1  converges to a neighborhood of the desired point with high probability. In other words, for any  $\epsilon, \beta > 0$, we determine the number of iterations $K$ required for any trajectory of Algorithm 1 to satisfy  
$\min_{k=1:K} \|\nabla F(\theta_k)\|^2 < \epsilon$ with probability at least \( 1 - \beta \). 

\begin{theorem}\label{th-ncvx-rate}
    Let \( \eta = \eta_0 K^{-1/4} \) and \( \gamma_k = \gamma_0 K^{-1/4} \). After \( K \) iterations of Algorithm 1, we have  
    \[
    \min_{k=1:K} \mathbb{E}[\|\nabla F(\theta_k)\|^2] \leq \frac{2\hat{\Delta}}{\sqrt{K} \eta_0 \gamma_0 c_1} + \frac{c_3^2\gamma_0^2}{\sqrt{K}} + \frac{\eta_0 \gamma_0 C \mu}{c_1 \sqrt{K}},
    \]
    indicating a convergence rate of order \( O(1/\sqrt{K}) \).

    Furthermore, if 
    \[
    K = \frac{1}{\epsilon^2\beta^2} \left( \frac{2\hat{\Delta}}{\eta_0 \gamma_0 c_1} + c_3^2 \gamma_0^2 + \frac{C \mu}{c_1} \eta_0 \gamma_0 \right)^2
    \]
    for any \( \epsilon, \beta > 0 \), then  
    \[
    \mathbb{P} \left( \min_{k=1:K} \|\nabla F(\theta_k)\|^2 < \epsilon \right) \geq 1 - \beta.
    \]
\end{theorem}

\noindent Proof: Refer to Appendix \ref{th-ncvx-rate-proof}.

The theorem means that, $\forall$ $\epsilon,\beta >0$, the algorithm achieves a neighborhood of $\|\nabla F(\theta_k)\|^2=0$, that is $min_{k=1:K} \|\nabla F(\theta_k)\|^2 < \epsilon$, after a number of iterations $K$ and with a probability $1-\beta$, where $\beta=\frac{1}{\epsilon \sqrt{K}}\biggl(\frac{2\hat{\Delta}}{\eta_0 \gamma_0 c_1}+c_3^2\gamma_0^2+\frac{c_2\mu}{ c_1} \eta_0\gamma_0\biggr)$. By taking $\epsilon <<1$ and $\beta <<1$, one can find the number of iterations K satisfying a sample path convergence to a neighborhood of the desired point with high probability $1-\beta$, as $K=\frac{1}{\epsilon^2\beta^2} \biggl(\frac{2\hat{\Delta}}{\eta_0 \gamma_0 c_1}+c_3^2\gamma_0^2+\frac{C\mu}{ c_1} \eta_0\gamma_0\biggr)^2$. 

\subsection{Convergence of Algorithm 2}
We provide now the convergence  of Algorithm 2. 
\begin{theorem}\label{th2-ncvx}
	When Assumptions \ref{objective_fct}-\ref{perturbation} hold, we have  $\lim_{k\rightarrow\infty}\mathbb{E}[\|\nabla F(\theta_k)\|^2]=0$  under Algorithm 2.
	
	Proof: Refer to Appendices B-A, B-B and \ref{th2-ncvx-proof}.
\end{theorem}
We then find the convergence rate, which turns out to have the same order as that of Algorithm 1, i.e.  $O(1/\sqrt{K})$. We also provide a high-probability sample path convergence result as for the previous algorithm.  For any $\epsilon, \beta >0$, we find the number of iterations $K$ required for any run of  algorithm 2 to achieve a neighborhood of $\|\nabla F(\theta_k)\|^2=0$, that is $min_{k=1:K} \|\nabla F(\theta_k)\|^2 < \epsilon$,  with  probability $1-\beta$. By taking $\epsilon, \beta << 1$, there exists a number of iterations $K$ requited to achieve  $min_{k=1:K} \|\nabla F(\theta_k)\|^2 < \epsilon$, i.e. $min_{k=1:K} \|\nabla F(\theta_k)\|^2 \approx 0 $, with probability $\approx 1$
These results are stated in the following theorem.  
Let $\eta = \eta_0(K)^{-1/2}$.
\begin{theorem}\label{th2-ncvx-rate}
	After $K$ iterations of Algorithm 1, we have    
	$\min_{k=1:K} \mathbb{E}[\|\nabla F(\theta_k)\|^2] \leq \frac{2\hat{\Delta}}{\sqrt{K}\eta_0 b_1}+\frac{\eta_0\mu C_2}{2b_1\sqrt{K}}$ meaning that the convergence rate is of order $O(1/\sqrt{K})$.\\ In addition, if $K=\frac{1}{\epsilon^2\beta^2} \biggl(\frac{\hat{\Delta}}{b_1\eta_0}+\frac{\eta_0\mu C_2}{2b_1} \biggr)^2$ for any $\epsilon,\beta>0$, we have 
$Prob\biggl(\min_{k=1:K} \|\nabla F(\theta_k)\|^2 < \epsilon\biggr)\geq 1-\beta$

	Proof: Refer to Appendix \ref{th-ncvx-rate-proof}.
\end{theorem}

\section{Asynchronous devices}

We consider in this section the case where the devices are asynchronous. Specifically, we assume that some devices transmit their first symbols $a_i$ at the same time when other devices are transmitting their gradient estimate (ZO or FO). At a first look, this interference between $a_i$ symbols and gradient estimates could be seen as destructive, hindering the system from tracking the accurate gradient descent. In this section, we prove that such an interference, and hence lack of synchronization,  cannot prevent the considered algorithms from converging to the desired point.   To illustrate that, let us consider that a subset of devices $\mathcal{N}_1$ transmits their first symbol $a$  when another subset $\mathcal{N}_2$ transmits the gradient estimates. We  first consider the ZO method and provide a modification of Algorithm 1 along with its convergence analysis, and then presents the modification and analysis of the first-order method. 

\subsection{Modified EZOFL}
 In this case, the server receives information in three slots. In the first slot, $\sum_{i \in \mathcal{N}_1} a_i h_{i,k} + n_{1,k}$ is received. In the second slot, $\sum_{i \in \mathcal{N}_1} \Delta f_{i,k} h_{i,k} +\sum_{j \in \mathcal{N}_2} a_j h_{j,k}+ n_{2,k}$ is received, while in the third slot, $\sum_{j \in \mathcal{N}_2} \Delta f_{j,k} h_{j,k} + n_{3,k}$
 is received. The server computes the following scalar and broadcast it to the devices: 
\begin{align}
  &   ( \sum_{i \in \mathcal{N}_1} a_i h_{i,k} + n_{1,k} )  ( \sum_{i \in \mathcal{N}_1} \Delta f_{i,k} h_{i,k} + \sum_{j \in \mathcal{N}_2} a_j h_{j,k}+ n_{2,k} )+ \nonumber \\ &  ( \sum_{j \in \mathcal{N}_2} \Delta f_{j,k} h_{j,k} + n_{3,k} )  ( \sum_{i \in \mathcal{N}_1} \Delta f_{i,k} h_{i,k} + \sum_{j \in \mathcal{N}_2} a_j h_{j,k}+ n_{2,k} ) \nonumber
\end{align}
Each device reconstructs the zero-order gradient estimate:
\begin{align}
\label{gradAsync}
    & \hat{g}_k= \Phi_k \times \biggl[ \nonumber \\ &   ( \sum_{i \in \mathcal{N}_1} a_i h_{i,k} + n_{1,k} )  ( \sum_{i \in \mathcal{N}_1} \Delta f_{i,k} h_{i,k} + \sum_{j \in \mathcal{N}_2} a_j h_{j,k}+ n_{2,k} )+ \nonumber \\ & ( \sum_{j \in \mathcal{N}_2} \Delta f_{j,k} h_{j,k} + n_{3,k} ) ( \sum_{i \in \mathcal{N}_1} \Delta f_{i,k} h_{i,k} + \sum_{j \in \mathcal{N}_2} a_j h_{j,k}+ n_{2,k} ) \biggr] \nonumber 
\end{align}
Finally, each device updates its model using:
\begin{equation}
    \theta_{k+1} = \theta_k - \eta_k \hat{g}_k
\end{equation}

We now provide the convergence of the algorithm  in the following theorem. 
\begin{theorem}\label{th-ncvx-async}
	When Assumptions \ref{objective_fct}-\ref{perturbation} hold, we have  $\lim_{k\rightarrow\infty}\mathbb{E}[\|\nabla F(\theta_k)\|^2]=0$.
    Furthermore, if 
    \[
    K = \frac{1}{\epsilon^2\beta^2} \left( \frac{2\hat{\Delta}}{\eta_0 \gamma_0 c_1} + c_3^2 \gamma_0^2 + \frac{C' \mu}{c_1} \eta_0 \gamma_0 \right)^2
    \]
    where $\hat{\Delta}=F(\theta_0)-F(\theta^*)$, then for any \( \epsilon, \beta > 0 \),
    \[
    \mathbb{P} \left( \min_{k=1:K} \|\nabla F(\theta_k)\|^2 < \epsilon \right) \geq 1 - \beta.
    \]
	
	Proof: The proof is provided in Appendix \ref{th-ncvx-proof-async}.
\end{theorem}
 
\subsection{Modified EFOFL}
In a way similar to the ZO case, the server receives information in three slots,  $\sum_{i \in \mathcal{N}_1} a_i h_{i,k} + n_{1,k}$, $\sum_{i \in \mathcal{N}_1} \left( \nabla f_{i,k}(\theta_k) \right)^T \Phi_k h_{i,k}  +\sum_{j \in \mathcal{N}_2} a_j h_{j,k}+ n_{2,k}$, and $\sum_{j \in \mathcal{N}_2} \left( \nabla f_{j,k}(\theta_k) \right)^T \Phi_k h_{j,k}  + n_{3,k}$. The server computes the following scalar and broadcast it to the devices: 
\begin{eqnarray}
    \label{prodAsync2}
  z_k &=  \biggl( \sum_{i \in \mathcal{N}_1} a_i h_{i,k} + n_{1,k} \biggr)  \biggl( \sum_{i \in \mathcal{N}_1} \left( \nabla f_{i,k}(\theta_k) \right)^T \Phi_k h_{i,k}  \nonumber  \\ &+\sum_{j \in \mathcal{N}_2} a_j h_{j,k}+ n_{2,k} \biggr) \nonumber \\ &+ \left( \sum_{j \in \mathcal{N}_2} \left( \nabla f_{j,k}(\theta_k) \right)^T \Phi_k h_{j,k} + n_{3,k} \right)\times \nonumber \\ &   \biggl( \sum_{i \in \mathcal{N}_1} \left( \nabla f_{i,k}(\theta_k) \right)^T \Phi_k  h_{i,k} + \sum_{j \in \mathcal{N}_2} a_j h_{j,k}+ n_{2,k} \biggr) \quad \nonumber
\end{eqnarray}
Each device reconstructs the gradient estimate:
\begin{eqnarray}
\label{gradAsync2}
    \tilde{g}_k &= \Phi_k \biggl[ ( \sum_{i \in \mathcal{N}_1} a_i h_{i,k} + n_{1,k} )  ( \sum_{i \in \mathcal{N}_1} \left( \nabla f_{i,k}(\theta_k) \right)^T \Phi_k h_{i,k} \nonumber \\ &+ \sum_{j \in \mathcal{N}_2} a_j h_{j,k}+ n_{2,k} ) \nonumber \\ &+ ( \sum_{j \in \mathcal{N}_2} \left( \nabla f_{j,k}(\theta_k) \right)^T \Phi_k  h_{j,k} + n_{3,k} ) \times \nonumber \\ & ( \sum_{i \in \mathcal{N}_1} \left( \nabla f_{i,k}(\theta_k) \right)^T \Phi_k h_{i,k}  + \sum_{j \in \mathcal{N}_2} a_j h_{j,k}+ n_{2,k} ) \biggr] \nonumber 
\end{eqnarray}
Finally, each device updates its model using:
\begin{equation}
    \theta_{k+1} = \theta_k - \eta_k \tilde{g}_k
\end{equation}

We now provide the convergence of the algorithm  in the following theorem. 
\begin{theorem}\label{th-ncvx-EFOFL-async}
	When Assumptions \ref{objective_fct}-\ref{perturbation} hold, we have  $\lim_{k\rightarrow\infty}\mathbb{E}[\|\nabla F(\theta_k)\|^2]=0$.
    Furthermore, if 
    \[
    K = \frac{1}{\epsilon^2\beta^2} \left( \frac{\hat{\Delta}}{b_1\eta_0} + \frac{C_2' \mu}{2b_1} \eta_0  \right)^2
    \]
    where $\hat{\Delta}=F(\theta_0)-F(\theta^*)$, then for any \( \epsilon, \beta > 0 \), 
    \[
    \mathbb{P} \left( \min_{k=1:K} \|\nabla F(\theta_k)\|^2 < \epsilon \right) \geq 1 - \beta.
    \]
	
	Proof: The proof is provided in Appendix \ref{th-ncvx-proof-async-EFOFL}.
\end{theorem}

\section{Numerical Results}\label{num}

\begin{figure}[h]
    \centering
    \includegraphics[width=0.9\linewidth]{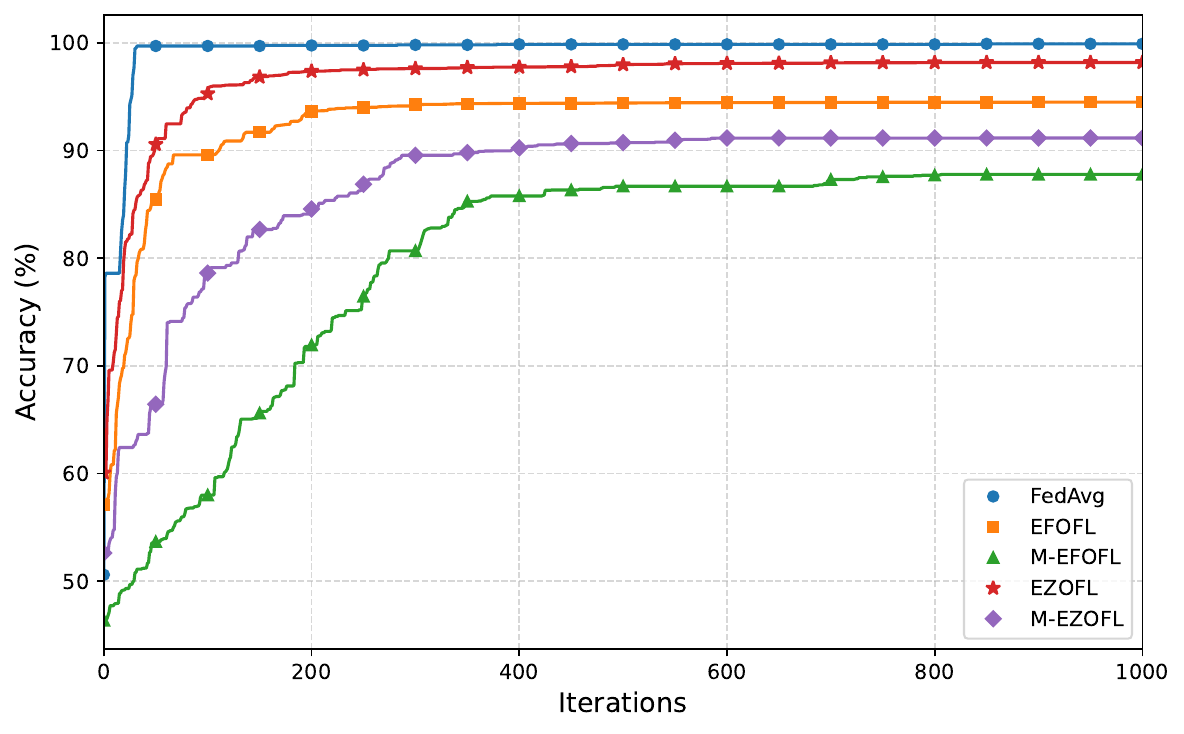}
    \caption{Test accuracy EZOFL and EFOFL  in comparison with  \textsc{FedAvg}.}
    \label{fig:accuracy_methods}
\end{figure}

\begin{figure}[h]
    \centering
    \includegraphics[width=0.9\linewidth]{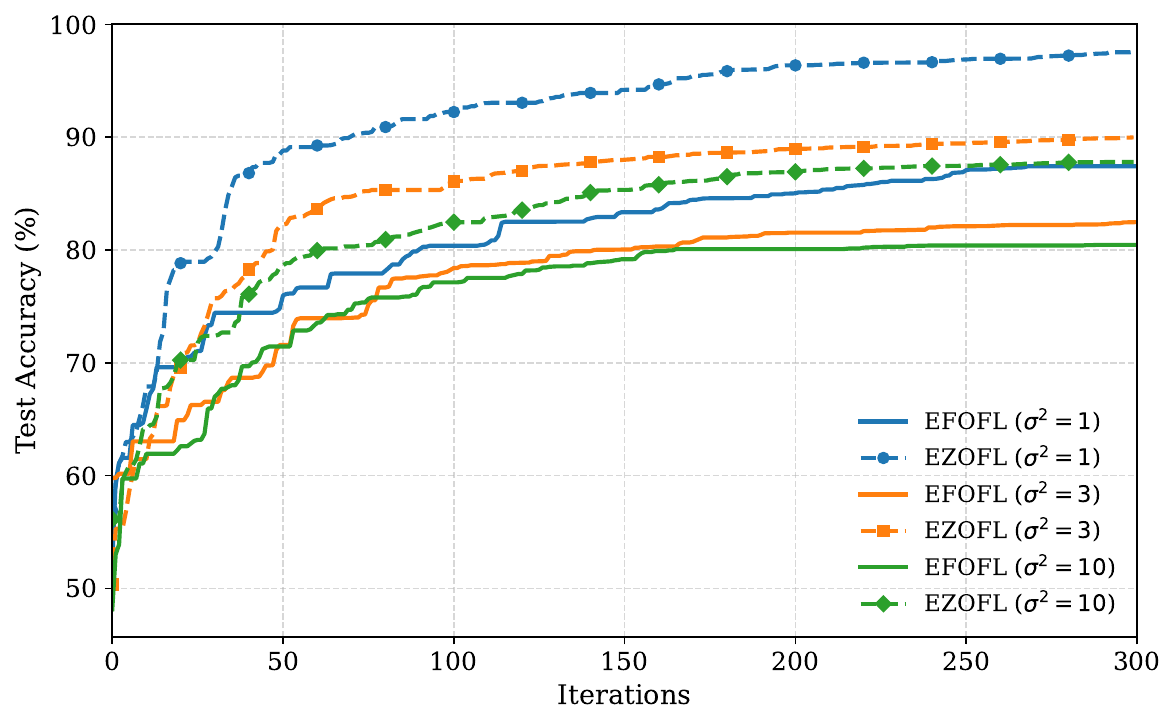}
    \caption{Test Accuracy  for EFOFL and EZOFL with different channel variances.}
    \label{fig:zofl_accuracy}
\end{figure}

We provide numerical results for both EZOFL and EFOFL algorithms over wireless networks. The performance is evaluated in a federated learning setup where devices communicate their updates under different channel conditions. The channel coefficient $h_i$  between each device and the server is generated using  gaussian distribution with mean zero and variance $\sigma_h^2$. 
We conduct our experiments using a binary classification task with the MNIST dataset. The data is divided equally between all participating device where each locally trains a non-convex logistic regression model. The images are preprocessed to have dimension $28 \times 28$ and are distributed among $N = 10$ devices. The step sizes in the algorithms are $\alpha_k = 0.5(1 + k)^{-0.50}$ and $\gamma_k = 2.5(1 + k)^{-0.25}$. Our results show that both algorithms demonstrate consistent performance across various random variations that influence each simulation, aligning well with our theoretical results. 
To assess robustness, we evaluate the achieved accuracy of both methods,  under different channel variances $\sigma_h^2$.  The results are shown in Figures~\ref{fig:accuracy_methods} and~\ref{fig:zofl_accuracy}, respectively, for EZOFL and EFOFL.

Both EFOFL and EZOFL achieve competitive performance in federated learning scenarios while significantly reducing communication overhead.  The numerical results show that EFOFL does not perform better than EZOFL, even though it involves gradient computation, due to the randomness of the perturbation vector and channel effects.

\section{Conclusion}
In this work, two communication-efficient federated learning (FL) methods over wireless networks are investigated: a zero-order (ZO) approach using a two-point gradient estimator and a randomized first-order (FO) method. By reducing communication to two scalars  per learning round and integrating channel effects into the learning process,  the need for explicit CSI estimation is eliminated.
We established theoretical convergence guarantees, demonstrating a rate of $O(1/\sqrt{K})$ in non-convex settings. The numerical results, using the MNIST dataset, validated the effectiveness of these methods, showing competitive accuracy with significantly reduced communication overhead.


\appendices

\section{Convergence of Algorithm 1} 

Contrary to the stochastic gradient that provides an unbiased estimate of the gradient,  expectation $\mathbb{E}(g_k)$ in our case represents a biased estimate, that is $\neq 0$, introducing challenges in establishing convergence.
\begin{equation}\label{2p_grdt_est2}
	\begin{split}
		g_k	= &\Phi_k \biggl(\sum_{i\in\mathcal{N}} a_i h_{i,k}+n_{1,k}\biggr)  \Bigg( \sum_{i\in\mathcal{N}} h_{i,k}\Delta f_{i,k} +n_{2,k}\Bigg) 
	\end{split}
\end{equation}
where $\Delta f_{i,k}=f_i\big(\theta_k + \gamma_k\Phi_k, \xi_{i,k}\big)-f_i\Big(\theta_k - \gamma_k\Phi_k, \xi_{i,k}\big)$.
The proof follows in several steps. We first analyze $\mathbb{E}(g_k)$ and $\mathbb{E}(g^2_k)$ in Lemmas \ref{biased_estimators} and \ref{norm}. We then prove in subsection A-C the convergence of the algorithm, i.e. Theorem \ref{th-ncvx}, and in subsection A-D the sample path convergence and convergence rate, i.e. Theorem \ref{th-ncvx-rate}.

Let $\mathcal{H}_k = \{\theta_0, \xi_0, \theta_1, \xi_1, ..., \theta_{k-1}, \xi_{k-1}, \theta_k\}$ denote the history sequence, then the following two Lemmas characterize our gradient estimates. 

\begin{lemma}\label{biased_estimators}
	Let Assumptions \ref{objective_fct}-\ref{perturbation} be satisfied and define the scalar value $c_1=2 \alpha_2$. Then, the gradient estimator is biased w.r.t. the objective function's exact gradient $\nabla F(\theta)$. Concretely, 		
	$$\mathbb{E}[g_k|\mathcal{H}_k] = c_1\gamma_k(\nabla F(\theta_k)+\delta_k),$$ 
	where $\delta_k$ is the bias term. In addition, there  is a scalar  $c_3>0$ such that $$\|\delta_k\| \leq c_3\gamma_k.$$
	
	Proof: Refer to Appendix \ref{app-grdt_est}.

	\begin{lemma}\label{norm}
		Let Assumptions \ref{objective_fct}-\ref{perturbation} hold. Then,
		$$\mathbb{E}[\|g_k\|^2|\mathcal{H}_k] \leq C$$ where\\ 
        $C=b_2^2 \biggl[\sigma_1^2\sigma_2^2 +4b_2^2L^2\gamma_k^2\sigma_1^2\sum_{i\in\mathcal{N}} \mathbb{E}(h^2_{i,k}) 
        +\sigma_2^2 \sum_{i\in\mathcal{N}} a^2_i \mathbb{E}(h^2_{i,k}) 
        +4b_2^2L^2\gamma_k^2\sum_{i\in\mathcal{N}}\sum_{j\in\mathcal{N}} a_i^2\mathbb{E}(h^2_{i,k})\mathbb{E}(h^2_{j,k}) \\
        +4b_2^2L^2\gamma_k^2\sum_{i\in\mathcal{N}}\sum_{j\in\mathcal{N}} a_i a_j\mathbb{E}(h^2_{i,k})\mathbb{E}(h^2_{j,k}) \biggr]$
		
		Proof: Refer to Appendix \ref{norm-sq}.
	\end{lemma}
\end{lemma}

\subsection{Proof of Lemma \ref{biased_estimators}: Biased Estimator}\label{app-grdt_est}

\allowdisplaybreaks  
\begin{align*}
    & \mathbb{E}[g_k|\mathcal{H}_k] = \mathbb{E} \Bigg[\Phi_k \biggl(\sum_{i\in\mathcal{N}} a_i h_{i,k} + n_{1,k} \biggr) \\
    &\quad \times \biggl( \sum_{i\in\mathcal{N}} h_{i,k} \Delta f_{i,k} + n_{2,k} \biggr) 
    \Bigg| \mathcal{H}_k \Bigg] \\
    &= \mathbb{E} \Bigg[ \Phi_k \Bigg( 
        \sum_{i,j\in\mathcal{N}} a_i h_{i,k} h_{j,k} \Delta f_{j,k} + n_{1,k} n_{2,k} \\
    &\quad + n_{1,k} \sum_{i\in\mathcal{N}} h_{i,k} \Delta f_{i,k} 
        + n_{2,k} \biggl( \sum_{i\in\mathcal{N}} a_i h_{i,k} + n_{1,k} \biggr) 
        \Bigg) \Bigg| \mathcal{H}_k \Bigg] \\
    &\overset{(a)}{=} \sum_{i=1}^{N} a_i \mathbb{E} (h^2_{i,k}) 
        \mathbb{E}_{\Phi_k,\xi_k} \bigg[ \Phi_k \Delta f_k \Big| \mathcal{H}_k \bigg] \\
    &= \sum_{i=1}^{N} a_i \mathbb{E} (h^2_{i,k}) 
        \mathbb{E}_{\Phi_k,\xi_k} \bigg[ \Phi_k \biggl( f_i(\theta_k + \gamma_k\Phi_k, \xi_{i,k}) \\
    &\quad - f_i(\theta_k - \gamma_k\Phi_k, \xi_{i,k}) \biggr) \Big| \mathcal{H}_k \bigg] \\
    &\overset{(b)}{=} \sum_{i=1}^{N} a_i \mathbb{E} (h^2_{i,k}) 
        \mathbb{E}_{\Phi_k} \bigg[ \Phi_k \biggl( F_i(\theta_k + \gamma_k\Phi_k) 
     - F_i(\theta_k - \gamma_k\Phi_k) \biggr) \Big| \mathcal{H}_k \bigg] \\
    &\overset{(c)}{=} \sum_{i=1}^{N} a_i\mathbb{E}(h^2_{i,k}) 
    \mathbb{E}_{\Phi_k} \bigg[ \Phi_k \Big[ F_i(\theta_k) 
    + \gamma_k\Phi_k^T\nabla F_i(\theta_k) \\
    &\quad + \frac{\gamma_k^2}{2}\Phi_k^T \nabla^2 F_i(\acute{\theta}_k)\Phi_k 
    - \big( F_i(\theta_k) - \gamma_k\Phi_k^T\nabla F_i(\theta_k) \\
    &\quad + \frac{\gamma_k^2}{2} \Phi_k^T \nabla^2 F_i(\grave{\theta}_k)\Phi_k \big) \Big] 
    \Big|\mathcal{H}_k \bigg] \\
    &= \sum_{i=1}^{N} a_i\mathbb{E}(h^2_{i,k}) 
    \mathbb{E}_{\Phi_k} \bigg[ \Phi_k \bigg( 2\gamma_k\Phi_k^T\nabla F_i(\theta_k) \\
    &\quad + \frac{\gamma_k^2}{2} \Phi_k^T (\nabla^2 F_i(\acute{\theta}_k) 
    - \nabla^2F_i(\grave{\theta}_k))\Phi_k \bigg) \Big|\mathcal{H}_k \bigg] \\
    &= 2\gamma_k \sum_{i=1}^{N} a_i\mathbb{E}(h^2_{i,k}) 
    \mathbb{E}_{\Phi_k} \big[\Phi_k \Phi_k^T \big|\mathcal{H}_k \big] \nabla F_i(\theta_k) \\
    &\quad + \frac{\gamma_k^2}{2} \sum_{i=1}^{N} a_i\mathbb{E}(h^2_{i,k}) 
    \mathbb{E}_{\Phi_k} \bigg[ \Phi_k\Phi_k^T (\nabla^2 F_i(\acute{\theta}_k) \\
    &\quad - \nabla^2F_i(\grave{\theta}_k))\Phi_k \Big|\mathcal{H}_k \bigg] 
    \overset{(d)}{=} 2b_1\gamma_k\sum_{i=1}^{N} a_i\mathbb{E}(h^2_{i,k}) \nabla F_i(\theta_k) \\ &+ \frac{\gamma_k^2}{2} \sum_{i=1}^{N} a_i   \mathbb{E}(h^2_{i,k}) 
    \mathbb{E}_{\Phi_k} \bigg[ \Phi_k\Phi_k^T (\nabla^2 F_i(\acute{\theta}_k) 
     - \nabla^2F_i(\grave{\theta}_k))\Phi_k \Big|\mathcal{H}_k \bigg] \\
    &\overset{(e)}{=} c_1 \gamma_k \bigg( \sum_{i=1}^{N} a_i\mathbb{E}(h^2_{i,k}) \nabla F_i(\theta_k) + \delta_k \bigg) \overset{(f)}{=} c_1 \gamma_k (\nabla F(\theta_k) + \delta_k). 
\end{align*}
where $(a)$ is due to fact that $h_{i,k}$ and $h_{j,k}$ are independent for $j\neq i$,   $\mathbb{E}(h_{i,k})=0$, $\mathbb{E}(n_{1,k})=\mathbb{E}(n_{2,k})=0$, and $h_{i,k}$, $n_{1,k}$, $n_{2,k}$ and $\Delta f_{i,k}$ are independent.  $(b)$ is by the definition in (\ref{F_i}), $(c)$ is by Taylor expansion and mean-valued theorem and considering $\acute{\theta}_k$ between $\theta_k$ and $\theta_k + \gamma_k\Phi_k$, and $\grave{\theta}_k$ between $\theta_k$ and $\theta_k -\gamma_k\Phi_k$. $(d)$ is due to Assumption \ref{perturbation}. In $(e)$,  we let $c_1=2 b_1 $ and $\delta_k = \frac{\gamma_k}{4 b_1}\sum_{i=1}^{N} a_i\mathbb{E}(h^2_{i,k})\mathbb{E}\bigg[\Phi_k\Phi_k^T (\nabla^2 F_i(\acute{\theta}_k)-\nabla^2F_i(\grave{\theta}_k))\Phi_k\Big|\mathcal{H}_k\bigg]$, and  in (f) we take $a_i=1/\mathbb{E}(h^2_{i,k})$.

\begin{equation*}
\begin{aligned}\label{2p_bias}
	\delta_k = \frac{\gamma_k}{4b_1}\sum_{i=1}^{N} a_i\mathbb{E}(h^2_{i,k})\mathbb{E}\bigg[\Phi_k\Phi_k^T (\nabla^2 F_i(\acute{\theta}_k)-\nabla^2F_i(\grave{\theta}_k))\Phi_k\Big|\mathcal{H}_k\bigg].
\end{aligned}
\end{equation*}
$\delta_k$ can be upper-bounded, using Assumptions \ref{perturbation} and \ref{objective_fct}, as
\begin{equation}
    \begin{aligned}
	\begin{split}
		&\|\delta_k\| 	\overset{(a)}{\leq}  \frac{\gamma_k}{4b_1}\sum_{i=1}^{N} a_i\mathbb{E}(h^2_{i,k})\mathbb{E}\bigg[\|\Phi_k\| \|\Phi_k^T\| \|\nabla^2 F_i(\acute{\theta}_k) \\ &-\nabla^2 F_i(\grave{\theta}_k)\|\|\Phi_k\|\Big|\mathcal{H}_k\bigg] 
		\overset{(b)}{\leq}  \frac{b b_2^3 N }{2b_1}\gamma_k 
		& \overset{(c)}{=}  c_3\gamma_k,\\
	\end{split}
    \end{aligned}
\end{equation}

where $(a)$ is due to Jensen's inequality, $(b)$ is due Assumptions \ref{perturbation} and \ref{objective_fct}, and in $(c)$, $c_3=\frac{b b_2^3 N }{2b_1}$.
\subsection{Proof of Lemma \ref{norm}: Expected Norm Squared of the Estimated Gradient}\label{norm-sq}
Bounding the norm squared of the gradient estimate,
\allowdisplaybreaks  
\begin{align*}
		&\mathbb{E}[\|g_k\|^2|\mathcal{H}_k]\\ 
		= &\mathbb{E}\bigg[\|\Phi_k (\sum_{i\in\mathcal{N}} a_i h_{i,k}+n_{1,k})  ( \sum_{i\in\mathcal{N}} h_{i,k}\Delta f_{i,k} +n_{2,k})\|^2\Big|\mathcal{H}_k\bigg]\\
		\overset{(a)}{\leq}  &\mathbb{E}\bigg[\|\Phi_k\|^2 \| (\sum_{i\in\mathcal{N}} a_i h_{i,k}+n_{1,k})  ( \sum_{i\in\mathcal{N}} h_{i,k}\Delta f_{i,k} +n_{2,k})\|^2\Big|\mathcal{H}_k\bigg]\\
        \overset{(b)}{\leq}   & b_2^2\mathbb{E}\bigg[ \biggl(\sum_{i\in\mathcal{N}} a_i h_{i,k}+n_{1,k}\biggr)^2  \Bigg( \sum_{i\in\mathcal{N}} h_{i,k}\Delta f_{i,k} +n_{2,k}\Bigg)^2 \Big|\mathcal{H}_k\bigg]\\
            \overset{(c)}{=}   & b_2^2\biggl[ \sigma_1^2\mathbb{E}\biggl[\bigg(\sum_{i\in\mathcal{N}} h_{i,k}\Delta f_{i,k}\biggr)^2 \Big|\mathcal{H}_k \biggr]+\sigma_2^2 \mathbb{E}\biggl[\biggl(\sum_{i\in\mathcal{N}} a_i h_{i,k}\biggr)^2\Big|\mathcal{H}_k\biggr]\\ &+\sigma_1^2\sigma_2^2+\mathbb{E}\biggl[\biggl(\sum_{i\in\mathcal{N}} a_i h_{i,k}\biggr)^2\bigg(\sum_{i\in\mathcal{N}} h_{i,k}\Delta f_{i,k}\biggr)^2\Big|\mathcal{H}_k\biggr] \\ &+ \mathbb{E}\biggl[2n_{1,k}\biggl(\sum_{i\in\mathcal{N}} a_i h_{i,k}\biggr)\Bigg( \sum_{i\in\mathcal{N}} h_{i,k}\Delta f_{i,k} +n_{2,k}\Bigg)^2\Big|\mathcal{H}_k\biggr] \\ &+ \mathbb{E}\biggl[2n_{2,k}\bigg(\sum_{i\in\mathcal{N}} h_{i,k}\Delta f_{i,k}\biggr)\biggl(\sum_{i\in\mathcal{N}} a_i h_{i,k}+n_{1,k}\biggr)^2\Big|\mathcal{H}_k\biggr] \\
           \overset{(d)}{=} &b_2^2 \biggl[\sigma_1^2\sigma_2^2 +\sigma_1^2\sum_{i\in\mathcal{N}} \mathbb{E}(h^2_{i,k})\mathbb{E}\biggl[(\Delta f_{i,k})^2\Big|\mathcal{H}_k\biggr] +\sigma_2^2 \sum_{i\in\mathcal{N}} a^2_i \mathbb{E}(h^2_{i,k}) \\ &+\mathbb{E}\biggl[\sum_{i\in\mathcal{N}}\sum_{j\in\mathcal{N}}\sum_{l\in\mathcal{N}}\sum_{m\in\mathcal{N}} a_i a_j h_{i,k}h_{j,k} h_{l,k}h_{m,k}\Delta f_{l,k}\Delta f_{m,k}\Big|\mathcal{H}_k\biggr] \biggr]\\
           \overset{(e)}{=} &b_2^2 \biggl[\sigma_1^2\sigma_2^2 +\sigma_1^2\sum_{i\in\mathcal{N}} \mathbb{E}(h^2_{i,k})\mathbb{E}\biggl[(\Delta f_{i,k})^2\Big|\mathcal{H}_k\biggr] +\sigma_2^2 \sum_{i\in\mathcal{N}} a^2_i \mathbb{E}(h^2_{i,k}) \\ &+\sum_{i\in\mathcal{N}}\sum_{j\in\mathcal{N}} a_i^2\mathbb{E}(h^2_{i,k})\mathbb{E}(h^2_{j,k})\mathbb{E}\biggl[(\Delta f_{i,k})^2\Big|\mathcal{H}_k\biggr] \\ &+\sum_{i\in\mathcal{N}}\sum_{j\in\mathcal{N}} a_i a_j\mathbb{E}(h^2_{i,k})\mathbb{E}(h^2_{j,k}) \mathbb{E}\biggl[\Delta f_{i,k}\Delta f_{j,k}\Big|\mathcal{H}_k\biggr] \biggr]\\
           \overset{(f)}{\leq} &b_2^2 \biggl[ \sigma_1^2\sum_{i\in\mathcal{N}} \mathbb{E}(h^2_{i,k})L^2 \mathbb{E}\biggl[\|2\gamma_k\Phi_k\|^2 \Big|\mathcal{H}_k\biggr] +\sigma_2^2 \sum_{i\in\mathcal{N}} a^2_i \mathbb{E}(h^2_{i,k}) \\ &+\sigma_1^2\sigma_2^2+\sum_{i\in\mathcal{N}}\sum_{j\in\mathcal{N}} a_i^2\mathbb{E}(h^2_{i,k})\mathbb{E}(h^2_{j,k})L^2\mathbb{E}\biggl[\|2\gamma_k\Phi_k\|^2 \\ &+\sum_{i\in\mathcal{N}}\sum_{j\in\mathcal{N}} a_i a_j\mathbb{E}(h^2_{i,k})\mathbb{E}(h^2_{j,k}) L^2\mathbb{E}\biggl[\|2\gamma_k\Phi_k\|^2\biggr]\biggr]\\
           \overset{(g)}{\leq} &b_2^2 \biggl[\sigma_1^2\sigma_2^2 +4b_2^2L^2\gamma_k^2\sigma_1^2\sum_{i\in\mathcal{N}} \mathbb{E}(h^2_{i,k})  +\sigma_2^2 \sum_{i\in\mathcal{N}} a^2_i \mathbb{E}(h^2_{i,k}) \\ &+4b_2^2L^2\gamma_k^2\sum_{i\in\mathcal{N}}\sum_{j\in\mathcal{N}} a_i^2\mathbb{E}(h^2_{i,k})\mathbb{E}(h^2_{j,k}) \\ &+4b_2^2L^2\gamma_k^2\sum_{i\in\mathcal{N}}\sum_{j\in\mathcal{N}} a_i a_j\mathbb{E}(h^2_{i,k})\mathbb{E}(h^2_{j,k}) \biggr] = C 
\end{align*}
where $(a)$ is due to Cauchy-Schwartz inequality, $(b)$ is due to Assumptions \ref{perturbation}  and that $\biggl(\sum_{i\in\mathcal{N}} a_i h_{i,k}+n_{1,k}\biggr)  \Bigg( \sum_{i\in\mathcal{N}} h_{i,k}\Delta f_{i,k} +n_{2,k}\Bigg)$ is a scalar, (c) is due to the fact that  $n_{1,k}$ and $n_{2,k}$ are i.i.d. and independent of other variables, (d) is due to the independence between $h_{i,k}$ and $\Delta f_{i,k}$ and to $\mathbb{E}(n_{1,k})=\mathbb{E}(n_{2,k})=0$, (e) is due the independence between $h_{i,k}$ and $\Delta f_{i,k}$ and  $\mathbb{E}(h_{i,k}h_{j,k})=0$ $\forall$ $i\neq j$, (f) is by Lipschitz continuity of $f_{i,k}$, that is $\|\Delta f_{i,k}\|=\|f_i\big(\theta_k + \gamma_k\Phi_k, \xi_{i,k}\big)-f_i\Big(\theta_k - \gamma_k\Phi_k, \xi_{i,k}\big)\| \leq L\|2\gamma_k \Phi_k\|$ where $L=\max_{\xi_{i,k}} L_{\xi_{i,k}}$, and (g) is due to Assumptions \ref{perturbation}.

\subsection{Proof of Theorem \ref{th-ncvx}: Convergence analysis}\label{th-ncvx-proof}
By $\mu$-smoothness inequality applied to function $F_i$, we have
\begin{equation*}
	\begin{split}
		F(\theta_{k+1})
		&\leq F(\theta_k)+\langle\nabla F(\theta_k), \theta_{k+1}-\theta_k\rangle +\frac{ \mu}{2}\|\theta_{k+1}-\theta_k\|^2.\\
	\end{split}
\end{equation*}
which implies,
\begin{equation}
	\begin{split}
		F(\theta_{k+1})
		&\leq F_i(\theta_k)-\eta_k\langle\nabla F(\theta_k), g_k\rangle +\frac{\eta_k^2 \mu}{2}\|g_k\|^2.\\
	\end{split}
\end{equation}
Taking the conditional expectation given $\mathcal{H}_k$,
\begin{align*}\label{ncvx-rate1}
		&\mathbb{E}[F(\theta_{K+1})|\mathcal{H}_K] \\
		\overset{(a)}{\leq} &F(\theta_k)-c_1\eta_k\gamma_k\langle\nabla F(\theta_k), \nabla F(\theta_k)+\delta_k\rangle +\frac{C\mu}{2}\eta_k^2 \gamma_k^2\\
		= &F(\theta_k)-c_1 \eta_k\gamma_k\|\nabla F(\theta_k)\|^2-c_1 \eta_k\gamma_k\langle\nabla F(\theta_k), \delta_k\rangle\\&+\frac{C\mu}{2}\eta_k^2 \gamma_k^2\\
		\overset{(b)}{\leq} &F(\theta_k)-c_1 \eta_k\gamma_k\|\nabla F(\theta_k)\|^2+\frac{c_1 \eta_k\gamma_k}{2}\|\nabla F(\theta_k)\|^2 \\&+\frac{c_1 \eta_k\gamma_k}{2}\|\delta_k\|^2+\frac{C\mu}{2}\eta_k^2 \gamma_k^2\\
		= &F(\theta_k)-\frac{c_1 \eta_k\gamma_k}{2}\|\nabla F(\theta_k)\|^2 +\frac{c_1 \eta_k\gamma_k}{2}\|\delta_k\|^2+\frac{C\mu}{2}\eta_k^2 \gamma_k^2\\
		\overset{(c)}{\leq} &F(\theta_k)-\frac{c_1 \eta_k\gamma_k}{2}\|\nabla F(\theta_k)\|^2 +\frac{c_1 c_3^2}{2}\eta_k\gamma_k^3+\frac{C\mu}{2}\eta_k^2 \gamma_k^2 
\end{align*}
where $(a)$ is by Lemmas \ref{biased_estimators} and \ref{norm}.
$(b)$ is due to $-\langle a,b\rangle\leq \frac{1}{2}\|a\|^2+\frac{1}{2}\|b\|^2$. $(c)$ is by Lemma \ref{norm}.

By considering the telescoping sum, we get 
\begin{align}
	\begin{split}
		& \mathbb{E}[F(\theta_{K+1})|\mathcal{H}_K]
		\leq F(\theta_0)-\frac{c_1}{2}\sum_{k=0}^{K}\eta_k\gamma_k\|\nabla F(\theta_k)\|^2 \\&+\frac{c_1 c_3^2}{2}\sum_{k=0}^{K}\eta_k\gamma_k^3+\frac{C\mu}{2}\sum_{k=0}^{K}\eta_k^2\gamma_k^2\\
		&\textit{implying,} 0 \leq  \Delta_k-\frac{c_1}{2}\sum_{k=0}^{K}\eta_k\gamma_k\|\nabla F(\theta_k)\|^2 \\& +\frac{c_1 c_3^2}{2}\sum_{k=0}^{K}\eta_k\gamma_k^3 +\frac{C\mu}{2}\sum_{k=0}^{K}\eta_k^2\gamma_k^2\\
	\end{split}
\end{align}
where $\Delta_k =F(\theta_0)-\mathbb{E}[F(\theta_{K+1})|\mathcal{H}_K]$.   By taking the expectation over all possible $\mathcal{H}_K$, we get, 
\begin{equation}
	\begin{split}
		&\sum_{k=0}^{K}\eta_k\gamma_k\mathbb{E}[\|\nabla F(\theta_k)\|^2]\\
		\leq &\frac{2}{c_1}\mathbb{E}[\Delta_k]+c_3^2\sum_{k=0}^{K}\eta_k\gamma_k^3+\frac{C\mu}{c_1}\sum_{k=0}^{K}\eta_k^2\gamma_k^2
	\end{split}\label{ineq-convergence}
\end{equation}
By Assumption \ref{step_sizes_1}, we have that $\mathbb{E}[\Delta_k]<\infty$. In addition, we know that $\lim_{K\rightarrow\infty}\sum_{k=0}^{K}\eta_k\gamma_k^3<\infty$ and $\lim_{K\rightarrow\infty}\sum_{k=0}^{K}\eta_k^2\gamma_k^2<\infty$. Hence, 
\begin{equation}\label{nablaF}
	\lim_{K\rightarrow\infty}\sum_{k=0}^{K}\eta_k\gamma_k\mathbb{E}[\|\nabla F(\theta_k)\|^2]	<\infty.
\end{equation}
Since $\sum_k\eta_k\gamma_k$ diverges by Assumption \ref{step_sizes_1}, we can show in a similar way as in \cite{ElissaJMLR24} that 
\begin{equation}
	\lim_{k\rightarrow\infty} \mathbb{E}[\|\nabla F(\theta_k)\|^2]=0.
\end{equation}

\subsection{Proof of Theorem \ref{th-ncvx-rate}: Sample path Convergence with high probability and convergences rate}\label{th-ncvx-rate-proof}
Let $\theta^*$ be the solution to the optimization problem in (1), that is $F(\theta^*) \leq F(\theta)$ $\forall \theta$.  From (\ref{ineq-convergence}), we have 
\begin{align*}\label{nablaF_2pt}
		&\sum_{k=0}^{K}\eta_k\gamma_k\mathbb{E}[\|\nabla F(\theta_k)\|^2]
		\leq &\frac{2}{c_1}\mathbb{E}[\Delta_k]+c_3^2\sum_{k=0}^{K}\eta_k\gamma_k^3+\frac{C\mu}{c_1}\sum_{k=0}^{K}\eta_k^2\gamma_k^2
\end{align*}
where $\Delta_k =F(\theta_0)-\mathbb{E}[F(\theta_{K+1})|\mathcal{H}_K]$. We have then $\Delta_k =F(\theta_0)-F(\theta^*)+F(\theta^*)-\mathbb{E}[F(\theta_{K+1})|\mathcal{H}_K] \leq F(\theta_0)-F(\theta^*)$.  This implies that 
\begin{equation*}\label{nablaF_2pt}
	\begin{split}
		&\sum_{k=0}^{K}\eta_k\gamma_k\mathbb{E}[\|\nabla F(\theta_k)\|^2]
		\leq \frac{2}{c_1}\hat{\Delta}+c_3^2\sum_{k=0}^{K}\eta_k\gamma_k^3+\frac{C\mu}{c_1}\sum_{k=0}^{K}\eta_k^2\gamma_k^2
	\end{split}
\end{equation*}
where $\hat{\Delta}=F(\theta_0)-F(\theta^*)$. Let $\eta_k=\eta$ and $\gamma_k=\gamma$, we have then 
 \begin{equation*}\label{nablaF_2pt}
	\begin{split}
		& K \min_{k=1:K}\eta \gamma \mathbb{E}[\|\nabla F(\theta_k)\|^2] \leq \sum_{k=0}^{K}\eta\gamma\mathbb{E}[\|\nabla F(\theta_k)\|^2]\\
		\leq &\frac{2}{c_1}\hat{\Delta}+c_3^2 K\eta\gamma^3+\frac{c_2\mu}{c_1} K\eta^2\gamma^2 \\
        & \textit{implying,} \min_{k=1:K} \mathbb{E}[\|\nabla F(\theta_k)\|^2] \leq \frac{2}{K\eta \gamma c_1}\hat{\Delta}+c_3^2\gamma^2+\frac{C\mu}{c_1} \eta\gamma
	\end{split}
\end{equation*}
Let $\eta = \eta_0(K)^{-1/4}$ and $\gamma_k = \gamma_0 (K)^{-1/4}$. One can see that the upper bound on $\min_{k=1:K} \mathbb{E}[\|\nabla F(\theta_k)\|^2] $ scales as $O(K^{-1/2})$. In addition, by using Markov inequality, we have 
\begin{align}
        &  Prob\biggl(\min_{k=1:K} \|\nabla F(\theta_k)\|^2 > \epsilon\biggr)  \leq \frac{\mathbb{E}\biggl(\min_{k=1:K} \|\nabla F(\theta_k)\|^2\biggr)}{\epsilon} \nonumber \\ & \leq \frac{\min_{k=1:K} \mathbb{E}[\|\nabla F(\theta_k)\|^2]}{\epsilon}  \nonumber  \leq \frac{2}{\epsilon K\eta \gamma c_1}\hat{\Delta}+\frac{c_3^2\gamma^2}{\epsilon}+\frac{c_2\mu}{\epsilon c_1} \eta\gamma \nonumber 
        \\ &=\frac{1}{\epsilon \sqrt{K}}\biggl(\frac{2\hat{\Delta}}{\eta_0 \gamma_0 c_1}+c_3^2\gamma_0^2+\frac{C\mu}{ c_1} \eta_0\gamma_0\biggr) \overset{\Delta}{=} \beta
\end{align}
Taking $\beta=\frac{1}{\epsilon \sqrt{K}}\biggl(\frac{2\hat{\Delta}}{\eta_0 \gamma_0 c_1}+c_3^2\gamma_0^2+\frac{C\mu}{ c_1} \eta_0\gamma_0\biggr)$ one can find $K=\frac{1}{\epsilon^2\beta^2} \biggl(\frac{2\hat{\Delta}}{\eta_0 \gamma_0 c_1}+c_3^2\gamma_0^2+\frac{C\mu}{ c_1} \eta_0\gamma_0\biggr)^2$. This concludes the proof. 

\section{Convergence of Algorithm 2}
The proof follows in several steps. Let $\mathcal{H}_k = \{\theta_0, \xi_0, \theta_1, \xi_1, ..., \theta_{k-1}, \xi_{k-1}, \theta_k\}$ denote the history sequence.  We first analyze $\mathbb{E}(\hat{g}_k|\mathcal{H}_k)$ and $\mathbb{E}(\hat{g}^2_k|\mathcal{H}_k)$. We then prove in subsection B-C the convergence of the algorithm, i.e. Theorem \ref{th2-ncvx}, and in subsection B-D the sample path convergence and convergence rate, i.e. Theorem \ref{th2-ncvx-rate}.

The following two Lemmas characterize the gradient estimates. 

\begin{lemma}\label{unbiased_estimators}
	Let Assumptions \ref{objective_fct}-\ref{perturbation} be satisfied. Then, 		
	$$\mathbb{E}[g'_k|\mathcal{H}_k] =b_1 \nabla F(\theta_k)$$ 
	
	Proof: Refer to Appendix \ref{app2-grdt_est}.
	
	\begin{lemma}\label{norm2}
		Let Assumptions \ref{objective_fct}-\ref{perturbation} hold. Then,
		$$\mathbb{E}[\|g'_k\|^2|\mathcal{H}_k] \leq C_2,$$ where
        $C_2=b_2^2 \biggl[\sigma_1^2\sigma_2^2$  $+b_2^2\sigma_1^2A^2\sum_{i\in\mathcal{N}} \mathbb{E}(h^2_{i,k})$   $+\sigma_2^2 \sum_{i\in\mathcal{N}} a^2_i \mathbb{E}(h^2_{i,k})  $ $+4b_2^2A^2\sum_{i\in\mathcal{N}}\sum_{j\in\mathcal{N}} a_i^2\mathbb{E}(h^2_{i,k})\mathbb{E}(h^2_{j,k})$  $+b_2^2A^2\sum_{i\in\mathcal{N}}\sum_{j\in\mathcal{N}} a_i a_j\mathbb{E}(h^2_{i,k})\mathbb{E}(h^2_{j,k}) \biggr]$
		
		Proof: Refer to Appendix \ref{norm-fo}.
	\end{lemma}
\end{lemma}

\subsection{Proof of Lemma \ref{unbiased_estimators}}\label{app2-grdt_est}
\allowdisplaybreaks  
\begin{align*}
		&\mathbb{E}[g'_k|\mathcal{H}_k] \\
		= &\mathbb{E}\Big[\Phi_k (\sum_{i\in\mathcal{N}} a_i h_{i,k}+n_{1,k}) ( \sum_{i\in\mathcal{N}} h_{i,k}(\nabla f_{i,k}(\theta_k))^T \Phi_k +n_{2,k})|\mathcal{H}_k\Big]\\
        = &\mathbb{E}\Big[\Phi_k \biggl(\biggl(\sum_{i\in\mathcal{N}}\sum_{j\in\mathcal{N}} a_i h_{i,k}h_{j,k}\biggl(\nabla f_{i,k}(\theta_k)\biggr)^T \Phi_k\biggr) +n_{1,k}n_{2,k} \\ &+n_{1,k} \Bigg( \sum_{i\in\mathcal{N}} h_{i,k}\biggl(\nabla f_{i,k}(\theta_k)\biggr)^T \Phi_k \Bigg) \\ &+n_{2,k}\biggl(\sum_{i\in\mathcal{N}} a_i h_{i,k}+n_{1,k}\biggr)\biggr)|\mathcal{H}_k\Big]\\
        \overset{(a)}{=} &\sum_{i=1}^N a_i\mathbb{E}(h^2_{i,k})\mathbb{E}_{\Phi_k,\xi_k}\Big[\Phi_k \biggl(\nabla f_{i,k}(\theta_k)\biggr)^T \Phi_k\Big|\mathcal{H}_k\Big]\\
        \overset{(b)}{=} &\sum_{i=1}^N a_i\mathbb{E}(h^2_{i,k})b_1\mathbb{E}_{\xi_k}\Big[ \biggl(\nabla f_{i,k}(\theta_k)\biggr)\Big|\mathcal{H}_k\Big]
         \overset{(c)}{=} b_1  \nabla F_{k}(\theta_k)
    \end{align*}
where $(a)$ is due to fact that $h_{i,k}$ and $h_{j,k}$ are independent for $j\neq i$,   $\mathbb{E}(h_{i,k})=0$, $\mathbb{E}(n_{1,k})=\mathbb{E}(n_{2,k})=0$, and $h_{i,k}$, $n_{1,k}$, $n_{2,k}$ and $\nabla f_{i,k}$ are independent.  $(b)$ is due to the fact that $\Phi_k$ is i.i.d., that is $\mathbb{E}\biggl(\Phi(l)\Phi(m)\biggr)=0$ $\forall$ $l\neq m$. (c) is due to the definition in (\ref{F_i}), $\nabla F_{i,k}(\theta_k)=\mathbb{E}_{\Phi_k,\xi_k}\Big[ \biggl(\nabla f_{i,k}(\theta_k)\biggr)\Big|\mathcal{H}_k\Big]$, and $a_i=1/\mathbb{E}(h^2_{i,k})$.

\subsection{Proof of Lemma \ref{norm2}: Expected Norm Squared of the Estimated Gradient}\label{norm-fo}
Bounding the norm squared of the gradient estimate,
\allowdisplaybreaks  
\begin{align*}
		&\mathbb{E}[\|g'_k\|^2|\mathcal{H}_k]\\ 
		= &\mathbb{E}\bigg[\|\Phi_k (\sum_{i\in\mathcal{N}} a_i h_{i,k}+n_{1,k})  ( \sum_{i\in\mathcal{N}} h_{i,k}(\nabla f_{i,k}(\theta_k))^T \Phi_k +n_{2,k})\|^2\Big|\mathcal{H}_k\bigg]\\
		\overset{(a)}{\leq}  &\mathbb{E}\bigg[\|\Phi_k\|^2 \| (\sum_{i\in\mathcal{N}} a_i h_{i,k}+n_{1,k}) \\ & \times ( \sum_{i\in\mathcal{N}} h_{i,k}(\nabla f_{i,k}(\theta_k))^T \Phi_k +n_{2,k})\|^2\Big|\mathcal{H}_k\bigg]\\
        \overset{(b)}{\leq}   & b_2^2\mathbb{E}\bigg[ (\sum_{i\in\mathcal{N}} a_i h_{i,k}+n_{1,k})^2  ( \sum_{i\in\mathcal{N}} h_{i,k}(\nabla f_{i,k}(\theta_k))^T \Phi_k +n_{2,k})^2 \Big|\mathcal{H}_k\bigg]\\
            \overset{(c)}{=}   & b_2^2\biggl[\sigma_1^2\sigma_2^2 +\sigma_1^2\mathbb{E}\biggl[\bigg(\sum_{i\in\mathcal{N}} h_{i,k}\biggl(\nabla f_{i,k}(\theta_k)\biggr)^T \Phi_k\biggr)^2 \Big|\mathcal{H}_k \biggr] \\ &+\sigma_2^2 \mathbb{E}\biggl[\biggl(\sum_{i\in\mathcal{N}} a_i h_{i,k}\biggr)^2\Big|\mathcal{H}_k\biggr]\\ &+\mathbb{E}\biggl[\biggl(\sum_{i\in\mathcal{N}} a_i h_{i,k}\biggr)^2\bigg(\sum_{i\in\mathcal{N}} h_{i,k}\biggl(\nabla f_{i,k}(\theta_k)\biggr)^T \Phi_k\biggr)^2\Big|\mathcal{H}_k\biggr] \\ &+ \mathbb{E}\biggl[2n_{1,k}(\sum_{i\in\mathcal{N}} a_i h_{i,k})( \sum_{i\in\mathcal{N}} h_{i,k}(\nabla f_{i,k}(\theta_k))^T \Phi_k \\ &+n_{2,k})^2\Big|\mathcal{H}_k\biggr] + \mathbb{E}\biggl[2n_{2,k}(\sum_{i\in\mathcal{N}} h_{i,k}(\nabla f_{i,k}(\theta_k))^T \Phi_k) \\ & \ \ \ \times (\sum_{i\in\mathcal{N}} a_i h_{i,k}+n_{1,k})^2\Big|\mathcal{H}_k\biggr] \\
           \overset{(d)}{=} &b_2^2 \biggl[\sigma_1^2\sigma_2^2 +\sigma_1^2\sum_{i\in\mathcal{N}} \mathbb{E}(h^2_{i,k})\mathbb{E}\biggl[\biggl(\biggl(\nabla f_{i,k}(\theta_k)\biggr)^T \Phi_k\biggr)^2\Big|\mathcal{H}_k\biggr] \\ &+\sigma_2^2 \sum_{i\in\mathcal{N}} a^2_i \mathbb{E}(h^2_{i,k}) +\mathbb{E}\biggl[\sum_{i\in\mathcal{N}}\sum_{j\in\mathcal{N}}\sum_{l\in\mathcal{N}}\sum_{m\in\mathcal{N}} a_i a_j h_{i,k}h_{j,k} \\ & \times h_{l,k}h_{m,k}\biggl(\nabla f_{l,k}(\theta_k)\biggr)^T \Phi_k \biggl(\nabla f_{m,k}(\theta_k)\biggr)^T \Phi_k\Big|\mathcal{H}_k\biggr] \biggr]\\
           \overset{(e)}{=} &b_2^2 \biggl[\sigma_1^2\sigma_2^2 +\sigma_1^2\sum_{i\in\mathcal{N}} \mathbb{E}(h^2_{i,k})\mathbb{E}\biggl[(\biggl(\nabla f_{i,k}(\theta_k)\biggr)^T \Phi_k)^2\Big|\mathcal{H}_k\biggr] \\ &+\sigma_2^2 \sum_{i\in\mathcal{N}} a^2_i \mathbb{E}(h^2_{i,k})+\sum_{i\in\mathcal{N}}\sum_{j\in\mathcal{N}} a_i a_j\mathbb{E}(h^2_{i,k})\mathbb{E}(h^2_{j,k}) \\ & \times \mathbb{E}\biggl[\biggl(\nabla f_{i,k}(\theta_k)\biggr)^T \Phi_k\biggl(\nabla f_{j,k}(\theta_k)\biggr)^T \Phi_k\Big|\mathcal{H}_k\biggr]\\ &+\sum_{i\in\mathcal{N}}\sum_{j\in\mathcal{N}} a_i^2\mathbb{E}(h^2_{i,k})\mathbb{E}(h^2_{j,k})\mathbb{E}\biggl[(\biggl(\nabla f_{j,k}(\theta_k)\biggr)^T \Phi_k)^2\Big|\mathcal{H}_k\biggr]\biggr] \\ 
           \overset{(f)}{\leq} &b_2^2 \biggl[\sigma_1^2\sigma_2^2 +b_2^2\sigma_1^2\sum_{i\in\mathcal{N}} \mathbb{E}(h^2_{i,k}) \mathbb{E}\biggl[\|\nabla f_{i,k}\|^2 \Big|\mathcal{H}_k\biggr] \\ &+\sigma_2^2 \sum_{i\in\mathcal{N}} a^2_i \mathbb{E}(h^2_{i,k}) \\ &+b_2^2\sum_{i\in\mathcal{N}}\sum_{j\in\mathcal{N}} a_i^2\mathbb{E}(h^2_{i,k})\mathbb{E}(h^2_{j,k})\mathbb{E}\biggl[\|\nabla f_{j,k}\|^2 \biggr]\\ &+b_2^2\sum_{i\in\mathcal{N}}\sum_{j\in\mathcal{N}} a_i a_j\mathbb{E}(h^2_{i,k})\mathbb{E}(h^2_{j,k}) \mathbb{E}\biggl[\|\|\nabla f_{i,k}\|\|\nabla f_{j,k}\|\biggr]\biggr]\\  
           \overset{(g)}{\leq} &b_2^2 \biggl[\sigma_1^2\sigma_2^2 +b_2^2\sigma_1^2A^2\sum_{i\in\mathcal{N}} \mathbb{E}(h^2_{i,k})  +\sigma_2^2 \sum_{i\in\mathcal{N}} a^2_i \mathbb{E}(h^2_{i,k}) \\ &+b_2^2A^2\sum_{i\in\mathcal{N}}\sum_{j\in\mathcal{N}} a_i^2\mathbb{E}(h^2_{i,k})\mathbb{E}(h^2_{j,k}) \\ &+b_2^2A^2\sum_{i\in\mathcal{N}}\sum_{j\in\mathcal{N}} a_i a_j\mathbb{E}(h^2_{i,k})\mathbb{E}(h^2_{j,k}) \biggr]
            = C_2
\end{align*}
    
where $(a)$ is due to Cauchy-Schwartz inequality, $(b)$ is due to Assumptions \ref{perturbation}  and that $\biggl(\sum_{i\in\mathcal{N}} a_i h_{i,k}+n_{1,k}\biggr)  \Bigg( \sum_{i\in\mathcal{N}} h_{i,k}\biggl(\nabla f_{j,k}(\theta_k)\biggr)^T \Phi_k +n_{2,k}\Bigg)$ is a scalar, (c) is due to the fact that  $n_{1,k}$ and $n_{2,k}$ are i.i.d. and independent of other variables, (d) is due to the independence between $h_{i,k}$ and $\nabla f_{i,k}$ and to $\mathbb{E}(n_{1,k})=\mathbb{E}(n_{2,k})=0$, (e) is due the independence between $h_{i,k}$ and $\nabla f_{i,k}$ and  $\mathbb{E}(h_{i,k}h_{j,k})=0$ $\forall$ $i\neq j$, (f) is by  $(\nabla f_{j,k}(\theta_k))^T \Phi_k \leq \|(\nabla f_{j,k}(\theta_k))^T \Phi_k\|=\|\nabla f_{j,k}(\theta_k)\|\| \Phi_k\|$ and $\| \Phi_k\| \leq b_2$ by Assumption \ref{perturbation}, and (g) is due to $\|\nabla f_{j,k}(\theta_k)\| \leq A$.

\subsection{Proof of Theorem \ref{th2-ncvx}: Convergence analysis}\label{th2-ncvx-proof}
By using  similar steps as in \cite{Ghadimi13} (since $\mathbb{E}[g'_k|\mathcal{H}_k]=\nabla F_{k}(\theta_k)$ and $\mathbb{E}[\|g'_k\|^2|\mathcal{H}_k] <\infty$), we can show  
\begin{equation}
	\begin{split}
		& \mathbb{E}[F(\theta_{K+1})|\mathcal{H}_K]
		\leq F(\theta_0)-b_1\sum_{k=0}^{K}\eta_k\|\nabla F(\theta_k)\|^2 +\frac{\mu C_2}{2}\sum_{k=0}^{K}\eta_k^2\\
		& \textit{implying that} \quad  0 \leq  \Delta_{k+1}-b_1\sum_{k=0}^{K}\eta_k\|\nabla F(\theta_k)\|^2  +\frac{\mu C_2}{2}\sum_{k=0}^{K}\eta_k^2\\
	\end{split}
\end{equation}
where $\Delta_k =F(\theta_0)-\mathbb{E}[F(\theta_{K+1})|\mathcal{H}_K]$.   By taking the expectation over all possible $\mathcal{H}_K$, we get, 
\begin{equation}
		b_1\sum_{k=0}^{K}\eta_k\mathbb{E}[\|\nabla F(\theta_k)\|^2]  
		\leq \mathbb{E}[\Delta_k]+\frac{\mu C_2}{2}\sum_{k=0}^{K}\eta_k^2
	\label{ineq2-convergence}
\end{equation}
By Assumption \ref{step_sizes_1}, we have that $\mathbb{E}[\Delta_k]<\infty$. In addition, we also know  that  $\lim_{K\rightarrow\infty}\sum_{k=0}^{K}\eta_k^2<\infty$. Hence, 
\begin{equation}\label{nablaF}
	\lim_{K\rightarrow\infty}\sum_{k=0}^{K}\eta_k\mathbb{E}[\|\nabla F(\theta_k)\|^2]	<\infty.
\end{equation}
Since $\sum_k\eta_k$ diverges by Assumption \ref{step_sizes_1}, we can show in a similar way as in \cite{ElissaJMLR24} that 
\begin{equation}
	\lim_{k\rightarrow\infty} \mathbb{E}[\|\nabla F(\theta_k)\|^2]=0.
\end{equation}

\subsection{Proof of Theorem \ref{th2-ncvx-proof}: Sample path Convergence with high probability and convergence rate}\label{th2-ncvx-rate-proof}
Let $\theta^*$ be the solution to the optimization problem in (1), that is $F(\theta^*) \leq F(\theta)$ $\forall \theta$.  From (\ref{ineq2-convergence}), we have 
\begin{equation}\label{nablaF_2pt}
		\sum_{k=0}^{K}\eta_k\mathbb{E}[\|\nabla F(\theta_k)\|^2]
		\leq \frac{\mathbb{E}[\Delta_k]}{b_1}+\frac{\mu C_2}{2b_1}\sum_{k=0}^{K}\eta_k^2
\end{equation}
where $\Delta_k =F(\theta_0)-\mathbb{E}[F(\theta_{K+1})|\mathcal{H}_K]$. We have then $\Delta_k =F(\theta_0)-F(\theta^*)+F(\theta^*)-\mathbb{E}[F(\theta_{K+1})|\mathcal{H}_K] \leq F(\theta_0)-F(\theta^*)$.  This implies that 
\begin{equation}\label{nablaF_2pt}
		\sum_{k=0}^{K}\eta_k\mathbb{E}[\|\nabla F(\theta_k)\|^2] 
		\leq  \frac{\hat{\Delta}}{b_1}+\frac{\mu C_2}{2b_1}\sum_{k=0}^{K}\eta_k^2
\end{equation}
where $\hat{\Delta}=F(\theta_0)-F(\theta^*)$. Let $\eta_k=\eta$, we have then 
 \begin{align*}\label{nablaF_2pt}
		& K \eta \min_{k=1:K}  \mathbb{E}[\|\nabla F(\theta_k)\|^2] \leq \eta \sum_{k=0}^{K}\mathbb{E}[\|\nabla F(\theta_k)\|^2]\\
		\leq &\frac{\hat{\Delta}}{b_1}+\frac{\mu C_2}{2b_1}\sum_{k=0}^{K}\eta_k^2 \\
        & \textit{implying that} \min_{k=1:K} \mathbb{E}[\|\nabla F(\theta_k)\|^2] \leq \frac{1}{K\eta b_1}\hat{\Delta}+\frac{\mu C_2}{2b_1} \eta
\end{align*}
Let $\eta = \eta_0(K)^{-1/2}$. One can see that the upper bound on $\min_{k=1:K} \mathbb{E}[\|\nabla F(\theta_k)\|^2] $ scales as $O(K^{-1/2})$. In addition, by using Markov inequality, we have 
\begin{align*}
        &  Prob\biggl(\min_{k=1:K} \|\nabla F(\theta_k)\|^2 > \epsilon\biggr)  \leq \frac{\mathbb{E}\biggl(\min_{k=1:K} \|\nabla F(\theta_k)\|^2\biggr)}{\epsilon} \\ & \leq \frac{\min_{k=1:K} \mathbb{E}[\|\nabla F(\theta_k)\|^2]}{\epsilon}  \\ & \leq \frac{1}{\epsilon K\eta b_1}\hat{\Delta}+\frac{\mu C_2}{2b_1\epsilon} \eta =\frac{1}{\epsilon \sqrt{K}}\biggl(\frac{\hat{\Delta}}{b_1\eta_0}+\frac{\mu C_2}{2b_1} \eta_0\biggr)
\end{align*}
Taking $\beta=\frac{1}{\epsilon \sqrt{K}}\biggl(\frac{\hat{\Delta}}{\eta_0b_1}+\frac{\mu C_2}{2b_1} \eta_0\biggr)$, one can find $K=\frac{1}{\epsilon^2\beta^2} \biggl(\frac{\hat{\Delta}}{b_1\eta_0}+\frac{\mu C_2}{2b_1} \eta_0\biggr)^2$. This concludes the proof. 

\section{Convergence of EZOFL for Asynchronous devices}\label{th-ncvx-proof-async}
The proof follows in three steps. We first prove that $\mathbb{E}[\hat{g}_k|\mathcal{H}_k]=c_1 \gamma_k(\nabla F(\theta_k)+\delta_k)$. We then prove that $\mathbb{E}[\|\hat{g}_k\|^2|\mathcal{H}_k]$ is bounded. By using these two results, and following a similar approach to the one in Appendix A, we prove the convergence. 

\allowdisplaybreaks  
1- \begin{align*}
		&\mathbb{E}[\hat{g}_k|\mathcal{H}_k] 
        = \mathbb{E}\Big[\Phi_k \biggl(\biggl(\sum_{i\in\mathcal{N}_1}\sum_{j\in\mathcal{N}_1} a_i h_{i,k}h_{j,k}\Delta f_{j,k}\biggr) +n_{1,k}n_{2,k}  \\ &+n_{1,k} \Bigg( \sum_{i\in\mathcal{N}_1} h_{i,k}\Delta f_{i,k} \Bigg)+n_{2,k}\biggl(\sum_{i\in\mathcal{N}_1} a_i h_{i,k}\biggr)\\&+ n_{1,k} \Bigg( \sum_{i\in\mathcal{N}_2} h_{i,k}a_{i} \Bigg)+\biggl(\sum_{i\in\mathcal{N}_2}\sum_{j\in\mathcal{N}_2} a_i h_{i,k}h_{j,k}\Delta f_{j,k}\biggr)  \\ &+ n_{2,k} \Bigg( \sum_{i\in\mathcal{N}_2} h_{i,k}\Delta f_{i,k} \Bigg)+n_{3,k}\biggl(\sum_{i\in\mathcal{N}_2} a_i h_{i,k}\biggr)+n_{3,k}n_{2,k} \\&+n_{3,k}  \sum_{i\in\mathcal{N}_1} h_{i,k}\Delta f_{i,k}+\sum_{i\in\mathcal{N}_1}\sum_{j\in\mathcal{N}_2}  h_{i,k}h_{j,k}\Delta f_{i,k}\Delta f_{j,k}\biggr)|\mathcal{H}_k\Big]\\
        \overset{(a)}{=} & \sum_{i \in \mathcal{N}_1} a_i\mathbb{E}(h^2_{i,k})\mathbb{E}_{\Phi_k,\xi_k}\Big[\Phi_k \Delta f_k\Big|\mathcal{H}_k\Big] \\ &+\sum_{i \in \mathcal{N}_2} a_i\mathbb{E}(h^2_{i,k})\mathbb{E}_{\Phi_k,\xi_k}\Big[\Phi_k \Delta f_k\Big|\mathcal{H}_k\Big]\\
        =& \sum_{i=1}^N a_i\mathbb{E}(h^2_{i,k})\mathbb{E}_{\Phi_k,\xi_k}\Big[\Phi_k \Delta f_k\Big|\mathcal{H}_k\Big]
		= \sum_{i=1}^{N} a_i\mathbb{E}(h^2_{i,k})\times \\ & \mathbb{E}_{\Phi_k,\xi_k}\bigg[\Phi_k\Big[f_i\big(\theta_k + \gamma_k\Phi_k, \xi_{i,k}\big) -f_i\Big(\theta_k - \gamma_k\Phi_k, \xi_{i,k}\big)\Big] \Big|\mathcal{H}_k\bigg]\\
		\overset{(b)}{=} &\sum_{i=1}^{N} a_i\mathbb{E}(h^2_{i,k})\mathbb{E}_{\Phi_k}\bigg[\Phi_k \Big[F_i\Big(\theta_k + \gamma_k\Phi_k\Big) \\
        &-F_i\Big(\theta_k - \gamma_k\Phi_k\Big)\Big]\Big|\mathcal{H}_k\bigg] \\
        \overset{(c)}{=} &2b_1\gamma_k\sum_{i=1}^{N} a_i\mathbb{E}(h^2_{i,k}) \nabla F_i(\theta_k)+\frac{\gamma_k^2}{2}\sum_{i=1}^{N} a_i\mathbb{E}(h^2_{i,k}) \\ &\times \mathbb{E}_{\Phi_k}\bigg[\Phi_k\Phi_k^T (\nabla^2 F_i(\acute{\theta}_k)-\nabla^2F_i(\grave{\theta}_k))\Phi_k\Big|\mathcal{H}_k\bigg]\\
		\overset{(d)}{=} &c_1 \gamma_k(\sum_{i=1}^{N} a_i\mathbb{E}(h^2_{i,k})\nabla F_i(\theta_k)+\delta_k)  \overset{(e)}{=}c_1 \gamma_k(\nabla F(\theta_k)+\delta_k)
        \end{align*}
where $(a)$ is due to fact that $h_{i,k}$ and $h_{j,k}$ are independent for $j\neq i$,   $\mathbb{E}(h_{i,k})=0$, $\mathbb{E}(n_{1,k})=\mathbb{E}(n_{2,k})=0$, and $h_{i,k}$, $n_{1,k}$, $n_{2,k}$ and $\Delta f_{i,k}$ are independent.  $(b)$ is by the definition in (\ref{F_i}), $(c)$ is by Taylor expansion and mean value theorem and considering $\acute{\theta}_k$ between $\theta_k$ and $\theta_k + \gamma_k\Phi_k$, and $\grave{\theta}_k$ between $\theta_k$ and $\theta_k -\gamma_k\Phi_k$. $(d)$ is due to Assumption \ref{perturbation}. In $(e)$,  we let $c_1=2 b_1 $ and in (f) we take $a_i=1/\mathbb{E}(h^2_{i,k})$.

Then, by using an analysis similar to that in Appendix A-A, one can easily show  that 
$\|\delta_k\| \leq c_3\gamma_k$ where $c_3=\frac{b b_2^3 N }{2b_1}$. \\

2- We then show that $\mathbb{E}[\|\hat{g}_k\|^2|\mathcal{H}_k]$ is bounded. 
\allowdisplaybreaks  
\begin{align*}
		&\mathbb{E}[\|\hat{g}_k\|^2|\mathcal{H}_k] 
		 = \mathbb{E}\bigg[\|\Phi_k \biggl(\sum_{i\in\mathcal{N}_1} a_i h_{i,k}+n_{1,k}\biggr) \\ &\times \Bigg( \sum_{i\in\mathcal{N}_1} h_{i,k}\Delta f_{i,k} +\sum_{j\in\mathcal{N}_2} a_j h_{j,k}  +n_{2,k}\Bigg)\|^2\Big|\mathcal{H}_k\bigg]\\
        &+ \mathbb{E}\bigg[\|\Phi_k \biggl(\sum_{j\in\mathcal{N}_2} \Delta f_{j,k}  h_{j,k}+n_{3,k}\biggr)  \\ & \times \Bigg( \sum_{i\in\mathcal{N}_1} h_{i,k}\Delta f_{i,k} +\sum_{j\in\mathcal{N}_2} a_j h_{j,k}+n_{2,k}\Bigg)\|^2\Big|\mathcal{H}_k\bigg]\\
        &+ 2\mathbb{E}\bigg[\Phi_k^2 \biggl(\sum_{i\in\mathcal{N}_1} a_i h_{i,k}+n_{1,k}\biggr) \biggl(\sum_{j\in\mathcal{N}_2} \Delta f_{j,k}  h_{j,k}+n_{3,k}\biggr) \times \\ \ & \ \ \ \Bigg( \sum_{i\in\mathcal{N}_1} h_{i,k}\Delta f_{i,k} +\sum_{j\in\mathcal{N}_2} a_j h_{j,k}+n_{2,k}\Bigg) \\ & \times  \Bigg( \sum_{i\in\mathcal{N}_1} h_{i,k}\Delta f_{i,k} +\sum_{j\in\mathcal{N}_2} a_j h_{j,k}+n_{2,k}\Bigg)\Big|\mathcal{H}_k\bigg]\\
        &=\mathbb{E}\bigg[\|\Phi_k (\sum_{i\in\mathcal{N}_1} a_i h_{i,k}+n_{1,k})  ( \sum_{i\in\mathcal{N}_1} h_{i,k}\Delta f_{i,k} +n_{2,k})\|^2\Big|\mathcal{H}_k\bigg]\\ 
        & + \mathbb{E}\bigg[\|\Phi_k \biggl(\sum_{i\in\mathcal{N}_1} a_i h_{i,k}+n_{1,k}\biggr)  \Bigg(  \sum_{j\in\mathcal{N}_2} a_j h_{j,k}\Bigg)\|^2\Big|\mathcal{H}_k\bigg]\\
        &+2\mathbb{E}\bigg[\Phi_k^2 \biggl(\sum_{i\in\mathcal{N}_1} a_i h_{i,k}+n_{1,k}\biggr)^2 \Bigg( \sum_{j\in\mathcal{N}_2} a_j h_{j,k}\Bigg)\\ &\times  \Bigg( \sum_{i\in\mathcal{N}_1} h_{i,k}\Delta f_{i,k} +n_{2,k}\Bigg)\Big|\mathcal{H}_k\bigg]\\ 
        &+ \mathbb{E}\bigg[\|\Phi_k \biggl(\sum_{j\in\mathcal{N}_2} \Delta f_{j,k}  h_{j,k}+n_{3,k}\biggr)  \Bigg( \sum_{i\in\mathcal{N}_1} h_{i,k}\Delta f_{i,k} \\ &+\sum_{j\in\mathcal{N}_2} a_j h_{j,k}+n_{2,k}\Bigg)\|^2\Big|\mathcal{H}_k\bigg]
        + 2\mathbb{E}\bigg[\Phi_k^2 \biggl(\sum_{i\in\mathcal{N}_1} a_i h_{i,k}+n_{1,k}\biggr) \\ &\times \Bigg( \sum_{i\in\mathcal{N}_1} h_{i,k}\Delta f_{i,k} +\sum_{j\in\mathcal{N}_2} a_j h_{j,k}+n_{2,k}\Bigg)  \biggl(\sum_{j\in\mathcal{N}_2} \Delta f_{j,k}  h_{j,k} \\ &+n_{3,k}\biggr)  \times \Bigg( \sum_{i\in\mathcal{N}_1} h_{i,k}\Delta f_{i,k} +\sum_{j\in\mathcal{N}_2} a_j h_{j,k}+n_{2,k}\Bigg)\Big|\mathcal{H}_k\bigg]
\end{align*}
One can see that, in the expression above and compared to the previous analysis in Appendix A.B, there are additional terms resulted from the  de-synchronization of the devices. We have shown in Appendix A.B that 
\begin{align*}
		 &\mathbb{E}\bigg[\|\Phi_k \biggl(\sum_{i\in\mathcal{N}_1} a_i h_{i,k}+n_{1,k}\biggr)  \Bigg( \sum_{i\in\mathcal{N}_1} h_{i,k}\Delta f_{i,k} +n_{2,k}\Bigg)\|^2\Big|\mathcal{H}_k\bigg]\\
           \leq &b_2^2 \biggl[\sigma_1^2\sigma_2^2 +4b_2^2L^2\gamma_k^2\sigma_1^2\sum_{i\in\mathcal{N}_1} \mathbb{E}(h^2_{i,k})  +\sigma_2^2 \sum_{i\in\mathcal{N}_1} a^2_i \mathbb{E}(h^2_{i,k}) \\ &+4b_2^2L^2\gamma_k^2\sum_{i\in\mathcal{N}_1}\sum_{j\in\mathcal{N}_1} a_i^2\mathbb{E}(h^2_{i,k})\mathbb{E}(h^2_{j,k}) \\ &+4b_2^2L^2\gamma_k^2\sum_{i\in\mathcal{N}}\sum_{j\in\mathcal{N}_1} a_i a_j\mathbb{E}(h^2_{i,k})\mathbb{E}(h^2_{j,k}) \biggr]
\end{align*} 
We can bound the remaining terms in a way similar to what is done in Appendix A-B. We skip the details for brevity  and we provide the bounds hereinafter. 
\allowdisplaybreaks  
\begin{align*}
		&\mathbb{E}[\|g_k\|^2|\mathcal{H}_k] 
		\leq b_2^2 \biggl[\sigma_1^2\sigma_2^2 +4b_2^2L^2\gamma_k^2\sigma_1^2\sum_{i\in\mathcal{N}_1} \mathbb{E}(h^2_{i,k}) \\ &+\sigma_2^2 \sum_{i\in\mathcal{N}_1} a^2_i \mathbb{E}(h^2_{i,k})  +4b_2^2L^2\gamma_k^2\sum_{i\in\mathcal{N}_1}\sum_{j\in\mathcal{N}_1} a_i^2\mathbb{E}(h^2_{i,k})\mathbb{E}(h^2_{j,k}) \\ &+4b_2^2L^2\gamma_k^2\sum_{i\in\mathcal{N}_1}\sum_{j\in\mathcal{N}_1} a_i a_j\mathbb{E}(h^2_{i,k})\mathbb{E}(h^2_{j,k}) +(\sum_{i\in\mathcal{N}_1} a^2_i \mathbb{E}(h^2_{i,k})+\sigma_1^2) \\ & \times \sum_{i\in\mathcal{N}_2} a^2_j \mathbb{E}(h^2_{j,k}) +4b_2^2L^2\gamma_k^2\sigma_2^2\sum_{i\in\mathcal{N}_2} \mathbb{E}(h^2_{i,k})  +\sigma_3^2 \sum_{i\in\mathcal{N}_2} a^2_i \mathbb{E}(h^2_{i,k}) \\ &+\sigma_3^2\sigma_2^2+4b_2^2L^2\gamma_k^2\sum_{i\in\mathcal{N}_2}\sum_{j\in\mathcal{N}_2} a_i^2\mathbb{E}(h^2_{i,k})\mathbb{E}(h^2_{j,k}) \\ &+4b_2^2L^2\gamma_k^2\sum_{i\in\mathcal{N}_2}\sum_{j\in\mathcal{N}_2} a_i a_j\mathbb{E}(h^2_{i,k})\mathbb{E}(h^2_{j,k}) \\ &+4b_2^2L^2\gamma_k^2\sigma_3^2 \sum_{i\in\mathcal{N}_1}  \mathbb{E}(h^2_{i,k})+16b_2^4L^4\gamma_k^4 \sum_{i\in\mathcal{N}_2}  \mathbb{E}(h^2_{i,k})  \sum_{i\in\mathcal{N}_1} \mathbb{E}(h^2_{i,k}) \\&+16b_2^2L^2\gamma_k^2\sum_{i\in\mathcal{N}_2} a_i \mathbb{E}(h^2_{i,k})  \sum_{i\in\mathcal{N}_1} a_i \mathbb{E}(h^2_{i,k})  \biggr]=C' 
\end{align*}
 One can see that the non-synchronization between the devices results in a higher upper bound as compared to the synchronous case. 

3- Then, we use the By $\mu$-smoothness inequality applied to function $F_i$, we have
\begin{align*}
	\begin{split}
		F(\theta_{k+1})
		&\leq F(\theta_k)+\langle\nabla F(\theta_k), \theta_{k+1}-\theta_k\rangle +\frac{ \mu}{2}\|\theta_{k+1}-\theta_k\|^2.
	\end{split}
\end{align*}
which implies,
\begin{align}
		F(\theta_{k+1})
		&\leq F_i(\theta_k)-\eta_k\langle\nabla F(\theta_k), g_k\rangle +\frac{\eta_k^2 \mu}{2}\|g_k\|^2
\end{align}
Taking the conditional expectation given $\mathcal{H}_k$,
\begin{align}
		&\mathbb{E}[F(\theta_{K+1})|\mathcal{H}_K] 
		\overset{(a)}{\leq} F(\theta_k)-c_1\eta_k\gamma_k\langle\nabla F(\theta_k), \nabla F(\theta_k)+\delta_k\rangle \nonumber \\ &+\frac{C'\mu}{2}\eta_k^2 \gamma_k^2 
		= F(\theta_k)-c_1 \eta_k\gamma_k\|\nabla F(\theta_k)\|^2 \nonumber \\ &-c_1 \eta_k\gamma_k\langle\nabla F(\theta_k), \delta_k\rangle +\frac{C'\mu}{2}\eta a_k^2 \gamma_k^2 \label{bound-grad-async}
\end{align}
We proceed in the same way as in Appendix A.C to show that 
\begin{align}
		&\sum_{k=0}^{K}\eta_k\gamma_k\mathbb{E}[\|\nabla F(\theta_k)\|^2] \nonumber \\
		\leq &\frac{2}{c_1}\mathbb{E}[\Delta_k]+c_3^2\sum_{k=0}^{K}\eta_k\gamma_k^3+\frac{C'\mu}{c_1}\sum_{k=0}^{K}\eta_k^2\gamma_k^2
\end{align}
where $\Delta_k =F(\theta_0)-\mathbb{E}[F(\theta_{K+1})|\mathcal{H}_K]$. One can see that the bound above is greater than that obtained in Appendix A.C, as $C'>C$. By Assumption \ref{step_sizes_1}, we have  $\mathbb{E}[\Delta_k]<\infty$. In addition, we know that $\lim_{K\rightarrow\infty}\sum_{k=0}^{K}\eta_k\gamma_k^3<\infty$ and $\lim_{K\rightarrow\infty}\sum_{k=0}^{K}\eta_k^2\gamma_k^2<\infty$. Hence, 
\begin{equation}\label{nablaF}
	\lim_{K\rightarrow\infty}\sum_{k=0}^{K}\eta_k\gamma_k\mathbb{E}[\|\nabla F(\theta_k)\|^2]	<\infty.
\end{equation}
Since $\sum_k\eta_k\gamma_k$ diverges by Assumption \ref{step_sizes_1}, we can show in a similar way as in \cite{ElissaJMLR24} that 
\begin{equation}
	\lim_{k\rightarrow\infty} \mathbb{E}[\|\nabla F(\theta_k)\|^2]=0.
\end{equation}
Since $C'>C$, the number of iterations required for convergence is higher. This can be shown in the following subsection. 

\subsection{Sample path Convergence with high probability and convergences rate}\label{th-ncvx-rate-proof-asynch}
Let $\theta^*$ be the solution to the optimization problem in (1), that is $F(\theta^*) \leq F(\theta)$ $\forall \theta$.  From (\ref{bound-grad-async}), we have 
\begin{equation}\label{nablaF_2pt}
	\begin{split}
		&\sum_{k=0}^{K}\eta_k\gamma_k\mathbb{E}[\|\nabla F(\theta_k)\|^2]\\
		\leq &\frac{2}{c_1}\mathbb{E}[\Delta_k]+c_3^2\sum_{k=0}^{K}\eta_k\gamma_k^3+\frac{C'\mu}{c_1}\sum_{k=0}^{K}\eta_k^2\gamma_k^2
	\end{split}
\end{equation}
where $\Delta_k =F(\theta_0)-\mathbb{E}[F(\theta_{K+1})|\mathcal{H}_K]$. 
By using similar step as in Appendix A.D, one can show 
\begin{equation}
	\begin{split}
        &  Prob\biggl(\min_{k=1:K} \|\nabla F(\theta_k)\|^2 > \epsilon\biggr)   \\ & \leq \frac{1}{\epsilon \sqrt{K}}\biggl(\frac{2\hat{\Delta}}{\eta_0 \gamma_0 c_1}+c_3^2\gamma_0^2+\frac{C'\mu}{ c_1} \eta_0\gamma_0\biggr)
	\end{split}
\end{equation}
where $\hat{\Delta}=F(\theta_0)-F(\theta^*)$. Taking $\beta=\frac{1}{\epsilon \sqrt{K}}\biggl(\frac{2\hat{\Delta}}{\eta_0 \gamma_0 c_1}+c_3^2\gamma_0^2+\frac{C'\mu}{ c_1} \eta_0\gamma_0\biggr)$, one can find $K=\frac{1}{\epsilon^2\beta^2} \biggl(\frac{2\hat{\Delta}}{\eta_0 \gamma_0 c_1}+c_3^2\gamma_0^2+\frac{C'\mu}{ c_1} \eta_0\gamma_0\biggr)^2$. This concludes the proof. 

 \section{Convergence of EFOFL for Asynchronous devices}\label{th-ncvx-proof-async-EFOFL}
The convergence follows in three steps. 

1- First, we show that $\mathbb{E}[\tilde{g}_k|\mathcal{H}_k]=\nabla F_{k}(\theta_k)$
\allowdisplaybreaks  
\begin{align*}
		&\mathbb{E}[\tilde{g}_k|\mathcal{H}_k] 
		= \mathbb{E}\Big[\Phi_k \biggl[ ( \sum_{i \in \mathcal{N}_1} a_i h_{i,k} + n_{1,k} )  \\ & \times ( \sum_{i \in \mathcal{N}_1} \left( \nabla f_{i,k}(\theta_k) \right)^T \Phi_k h_{i,k}  + \sum_{j \in \mathcal{N}_2} a_j h_{j,k}+ n_{2,k} ) \nonumber \\ &+ ( \sum_{j \in \mathcal{N}_2} \left( \nabla f_{j,k}(\theta_k) \right)^T \Phi_k  h_{j,k} + n_{3,k} ) \times \nonumber \\ & ( \sum_{i \in \mathcal{N}_1} \left( \nabla f_{i,k}(\theta_k) \right)^T \Phi_k h_{i,k}  + \sum_{j \in \mathcal{N}_2} a_j h_{j,k}+ n_{2,k} ) \biggr]|\mathcal{H}_k\Big]\\
        \overset{(a)}{=} &\sum_{i \in \mathcal{N}_1} a_i\mathbb{E}(h^2_{i,k})\mathbb{E}_{\Phi_k,\xi_k}\Big[\Phi_k \biggl(\nabla f_{i,k}(\theta_k)\biggr)^T \Phi_k\Big|\mathcal{H}_k\Big] \nonumber \\ &+ \sum_{i \in \mathcal{N}_2} a_i\mathbb{E}(h^2_{i,k})\mathbb{E}_{\Phi_k,\xi_k}\Big[\Phi_k \biggl(\nabla f_{i,k}(\theta_k)\biggr)^T \Phi_k\Big|\mathcal{H}_k\Big] \nonumber \\
        \overset{(b)}{=} &\sum_{i=1}^N a_i\mathbb{E}(h^2_{i,k})b_1\mathbb{E}_{\xi_k}\Big[ \biggl(\nabla f_{i,k}(\theta_k)\biggr)\Big|\mathcal{H}_k\Big]
         \overset{(c)}{=}  b_1 \nabla F_{k}(\theta_k)
    \end{align*}
where $(a)$ is due to fact that $h_{i,k}$ and $h_{j,k}$ are independent for $j\neq i$,   $\mathbb{E}(h_{i,k})=0$, $\mathbb{E}(n_{1,k})=\mathbb{E}(n_{2,k})=0$, and $h_{i,k}$, $n_{1,k}$, $n_{2,k}$ and $\nabla f_{i,k}$ are independent.  $(b)$ is due to the fact that $\Phi_k$ is i.i.d., that is $\mathbb{E}\biggl(\Phi(l)\Phi(m)\biggr)=0$ $\forall$ $l\neq m$. (c) is due to the definition in (\ref{F_i}), $\nabla F_{i,k}(\theta_k)=\mathbb{E}_{\xi_k}\Big[ \biggl(\nabla f_{i,k}(\theta_k)\biggr)\Big|\mathcal{H}_k\Big]$, and $a_i=1/\mathbb{E}(h^2_{i,k})$.

2- We then show that $\mathbb{E}[\|\tilde{g}_k\|^2|\mathcal{H}_k]$ is bounded. 
\allowdisplaybreaks  
\begin{align*}
		&\mathbb{E}[\|\tilde{g}_k\|^2|\mathcal{H}_k]
		 = \mathbb{E}\biggl[\|\Phi_k \biggl(\sum_{i\in\mathcal{N}_1} a_i h_{i,k}+n_{1,k}\biggr) \times \\ & \biggl( \sum_{i\in\mathcal{N}_1} \left(\nabla f_{i,k}(\theta_k) \right)^T \Phi_k h_{i,k} +\sum_{j\in\mathcal{N}_2} a_j h_{j,k}+n_{2,k}\biggr)\|^2\Big|\mathcal{H}_k\biggr]\\
        &+ \mathbb{E}\biggl[\|\Phi_k \biggl(\sum_{j\in\mathcal{N}_2} \left(\nabla f_{j,k}(\theta_k) \right)^T \Phi_k  h_{j,k}+n_{3,k}\biggr) \times \\ & \biggl( \sum_{i\in\mathcal{N}_1} \left(\nabla f_{i,k}(\theta_k) \right)^T \Phi_k h_{i,k} +\sum_{j\in\mathcal{N}_2} a_j h_{j,k}+n_{2,k}\biggr)\|^2\Big|\mathcal{H}_k\biggr]\\
        &+ 2\mathbb{E}\biggl[\Phi_k^2 \biggl(\sum_{i\in\mathcal{N}_1} a_i h_{i,k}+n_{1,k}\biggr) \times \\&  \biggl(\sum_{i\in\mathcal{N}_1} \left(\nabla f_{i,k}(\theta_k) \right)^T \Phi_k h_{i,k} +\sum_{j\in\mathcal{N}_2} a_j h_{j,k}+n_{2,k}\biggr) \times \\ & \biggl(\sum_{j\in\mathcal{N}_2} \left(\nabla f_{j,k}(\theta_k) \right)^T \Phi_k  h_{j,k}+n_{3,k}\biggr) \times \\ & \biggl( \sum_{i\in\mathcal{N}_1} \left(\nabla f_{i,k}(\theta_k) \right)^T \Phi_k h_{i,k} +\sum_{j\in\mathcal{N}_2} a_j h_{j,k}+n_{2,k}\biggr)\Big|\mathcal{H}_k\biggr]\\
         \leq &b_2^2 \biggl[\sigma_1^2\sigma_2^2 +b_2^2\sigma_1^2A^2\sum_{i\in\mathcal{N}_1} \mathbb{E}(h^2_{i,k})  +\sigma_2^2 \sum_{i\in\mathcal{N}_1} a^2_i \mathbb{E}(h^2_{i,k}) \\ &+b_2^2A^2\sum_{i\in\mathcal{N}_1}\sum_{j\in\mathcal{N}_1} a_i^2\mathbb{E}(h^2_{i,k})\mathbb{E}(h^2_{j,k}) \\ &+b_2^2A^2\sum_{i\in\mathcal{N}_1}\sum_{j\in\mathcal{N}_1} a_i a_j\mathbb{E}(h^2_{i,k})\mathbb{E}(h^2_{j,k}) \\
         &+(\sum_{i\in\mathcal{N}_1} a^2_i \mathbb{E}(h^2_{i,k})+\sigma_1^2)\sum_{j\in\mathcal{N}_2} a^2_j \mathbb{E}(h^2_{j,k})\\ &+\sigma_3^2\sigma_2^2 +b_2^2A^2\sigma_2^2\sum_{i\in\mathcal{N}_2} \mathbb{E}(h^2_{i,k})  +\sigma_3^2 \sum_{i\in\mathcal{N}_2} a^2_i \mathbb{E}(h^2_{i,k}) \\ &+b_2^2A^2\sum_{i\in\mathcal{N}_2}\sum_{j\in\mathcal{N}_2} a_i^2\mathbb{E}(h^2_{i,k})\mathbb{E}(h^2_{j,k}) \\  &+b_2^2A^2\sum_{i\in\mathcal{N}_2}\sum_{j\in\mathcal{N}_2} a_i a_j\mathbb{E}(h^2_{i,k})\mathbb{E}(h^2_{j,k}) \\ &+b_2^2A^2\sigma_3^2 \sum_{i\in\mathcal{N}_2}  \mathbb{E}(h^2_{i,k})+b_2^4A^4 \sum_{i\in\mathcal{N}_1}\sum_{j\in\mathcal{N}_2}  \mathbb{E}(h^2_{i,k})    \mathbb{E}(h^2_{j,k}) \\&+4b_2^2A^2\sum_{i\in\mathcal{N}_1}\sum_{j\in\mathcal{N}_2} a_i a_j \mathbb{E}(h^2_{i,k})   \mathbb{E}(h^2_{i,k})  \biggr]=C_2'
\end{align*}
 where the upper bound is obtained by using similar steps as in Appendices A, B and C. 

 3- We then use the $\mu$-smoothness of the nonconvex objective 
 \begin{align*}
		F(\theta_{k+1})
		&\leq F(\theta_k)+\langle\nabla F(\theta_k), \theta_{k+1}-\theta_k\rangle +\frac{ \mu}{2}\|\theta_{k+1}-\theta_k\|^2
\end{align*}
which implies, using similar steps as in Appendix B-C, 
\begin{align}\label{nablaF_2pt-2}
		\sum_{k=0}^{K}\eta_k\mathbb{E}[\|\nabla F(\theta_k)\|^2] 
		\leq \frac{\Delta_k}{b_1}+\frac{\mu C_2'}{2b_1}\sum_{k=0}^{K}\eta_k^2
\end{align}
where $\Delta_k =F(\theta_0)-\mathbb{E}[F(\theta_{K+1})|\mathcal{H}_K]$, which in turn implies 
\begin{equation}\label{nablaF_2pt-2}
		\lim_{K \rightarrow \infty} \sum_{k=0}^{K}\eta_k\mathbb{E}[\|\nabla F(\theta_k)\|^2]		\leq \infty
\end{equation}
which also implies, using similar analysis to \cite{ElissaJMLR24}, that
\begin{equation}
	\lim_{k\rightarrow\infty} \mathbb{E}[\|\nabla F(\theta_k)\|^2]=0.
\end{equation}

4- Then, by following similar steps as in Appendix B.D, 
\begin{align*}\label{nablaF_2pt}
		& K \eta \min_{k=1:K}  \mathbb{E}[\|\nabla F(\theta_k)\|^2] \leq \eta \sum_{k=0}^{K}\mathbb{E}[\|\nabla F(\theta_k)\|^2]\\
		\leq &\frac{\hat{\Delta}}{b_1}+\frac{\mu C_2'}{2b_1}\sum_{k=0}^{K}\eta_k^2 \\
        & \textit{implying that} \min_{k=1:K} \mathbb{E}[\|\nabla F(\theta_k)\|^2] \leq \frac{1}{K\eta b_1}\hat{\Delta}+\frac{\mu C_2}{2b_1} \eta
\end{align*}
Let $\eta = \eta_0(K)^{-1/2}$. By using Markov inequality, we have 
\begin{align*}
        &  Prob\biggl(\min_{k=1:K} \|\nabla F(\theta_k)\|^2 > \epsilon\biggr)  \leq \frac{\mathbb{E}\biggl(\min_{k=1:K} \|\nabla F(\theta_k)\|^2\biggr)}{\epsilon} \\ & \leq \frac{\min_{k=1:K} \mathbb{E}[\|\nabla F(\theta_k)\|^2]}{\epsilon}   \leq \frac{1}{\epsilon \sqrt{K}}\biggl(\frac{\hat{\Delta}}{b_1\eta_0}+\frac{\mu C_2'}{2b_1} \eta_0\biggr) \overset{\Delta}{=} \beta
\end{align*}
which implies that $K=\frac{1}{\epsilon^2\beta^2} \left( \frac{\hat{\Delta}}{b_1\eta_0} + \frac{C_2' \mu}{2b_1} \eta_0  \right)^2$
\bibliography{example_paper}
\bibliographystyle{IEEEtran}

\end{document}